\documentclass[lettersize,journal]{IEEEtran}
\usepackage{amsmath,amsfonts}
\usepackage{amssymb}

\usepackage{algorithm}
\usepackage{algpseudocode}
\usepackage{array}
\usepackage[caption=false,font=normalsize,labelfont=sf,textfont=sf]{subfig}
\usepackage{textcomp}
\usepackage{stfloats}
\usepackage{url}
\usepackage{verbatim}
\usepackage{graphicx}
\usepackage{xcolor} 
\usepackage{adjustbox}
\usepackage{hhline}
\usepackage{colortbl}
\usepackage{tabularray}
\usepackage{multirow}
\usepackage{threeparttable}
\usepackage{wrapfig}
\usepackage{physics}
\usepackage{hyperref}
\usepackage{booktabs}
\usepackage{graphicx}       
\usepackage{cite} 
\usepackage{float}

\begin{document}

\title{Mobile-4DGS: Unified Static-Dynamic Real-time Mobile Gaussian Splatting}

\author{
\IEEEcompsocitemizethanks{
This research is funded in part by ARC-Discovery grant (DP220100800 to XY), ARC-DECRA grant (DE230100477 to XY), NVIDIA academic grant program, and Google Research Scholar Program.
 We thank all anonymous reviewers and editors for their constructive suggestions.}
  Xiaobiao Du \href{https://orcid.org/0000-0002-0085-962X}{\includegraphics[scale=0.5]{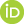}} \space 
   Zhen Fang \href{https://orcid.org/0000-0003-0602-6255}{\includegraphics[scale=0.5]{bio/ORCID.png}} \space 
   Tianqing Zhu \href{https://orcid.org/0000-0003-3411-7947} {\includegraphics[scale=0.5]{bio/ORCID.png}} \space   
Richard Hartley \href{https://orcid.org/0000-0002-5005-0191}
{\includegraphics[scale=0.5]{bio/ORCID.png}}~\IEEEmembership{IEEE Fellow} \space 
  Xin Yu \href{https://orcid.org/0000-0002-0269-5649}{\includegraphics[scale=0.5]{bio/ORCID.png}}~\IEEEmembership{Senior~Member}
 \thanks{Xiaobiao Du and Zhen Fang are with the University of Technology Sydney, Sydney 2009, Australia.} 
\thanks {Tianqing Zhu is with City University of Macau, Macau, 999078, China.}
\thanks {Richard Hartley is with the Australian National University, Canberra, 2600, Australia.}
 \thanks{Xin Yu is with Adelaide University, Adelaide, 5005, Australia. (email: xin.yu@adelaide.edu.au)}
}

\markboth{Journal of \LaTeX\ Class Files,~Vol.~14, No.~8, August~2026}%
{Shell \MakeLowercase{\textit{et al.}}: A Sample Article Using IEEEtran.cls for IEEE Journals}


\maketitle

\begin{abstract}
Recent advances in 3D Gaussian Splatting (3DGS) have achieved remarkable performance in novel view synthesis, yet deploying both static and dynamic Gaussian representations on resource-constrained mobile devices remains challenging due to heavy storage, redundant primitives, and costly per-frame computation. We present Mobile-4DGS, a unified lightweight framework for high-fidelity real-time static and dynamic Gaussian rendering on mobile platforms. For compact appearance modeling, we introduce a Monte Carlo Specular Energy Aggregator that compresses high-order radiance residuals into the first-order Spherical Harmonics (SH), together with an Attribute-Conditioned SH Enhancement module whose predicted offsets are pre-baked before inference. We further propose a Multi-View Alpha-Based Densification and Pruning strategy to suppress redundant primitives while maintaining multi-view consistency. For dynamic scenes, we develop a compact explicit 4D representation by constructing second-order Gaussian motion, learnable temporal support, and a binary static-dynamic partition, enabling continuous-time modeling without runtime deformation networks. Based on this partition, a Depth-Order Certificate selectively reuses previously committed depth orders to reduce re-projection, sorting, merging, and index-buffer updates during playback. Extensive experiments on static and dynamic scenes demonstrate that Mobile-4DGS substantially reduces storage and rendering overhead while maintaining competitive visual quality, enabling real-time 3D and 4D Gaussian Splatting on mobile devices. \textcolor{magenta}{\href{https://xiaobiaodu.github.io/mobile-4dgs-project/}{Code has been released: https://xiaobiaodu.github.io/mobile-4dgs-project/}}.
\end{abstract}

\begin{figure}

    \centering
    \includegraphics[width=0.99\linewidth]{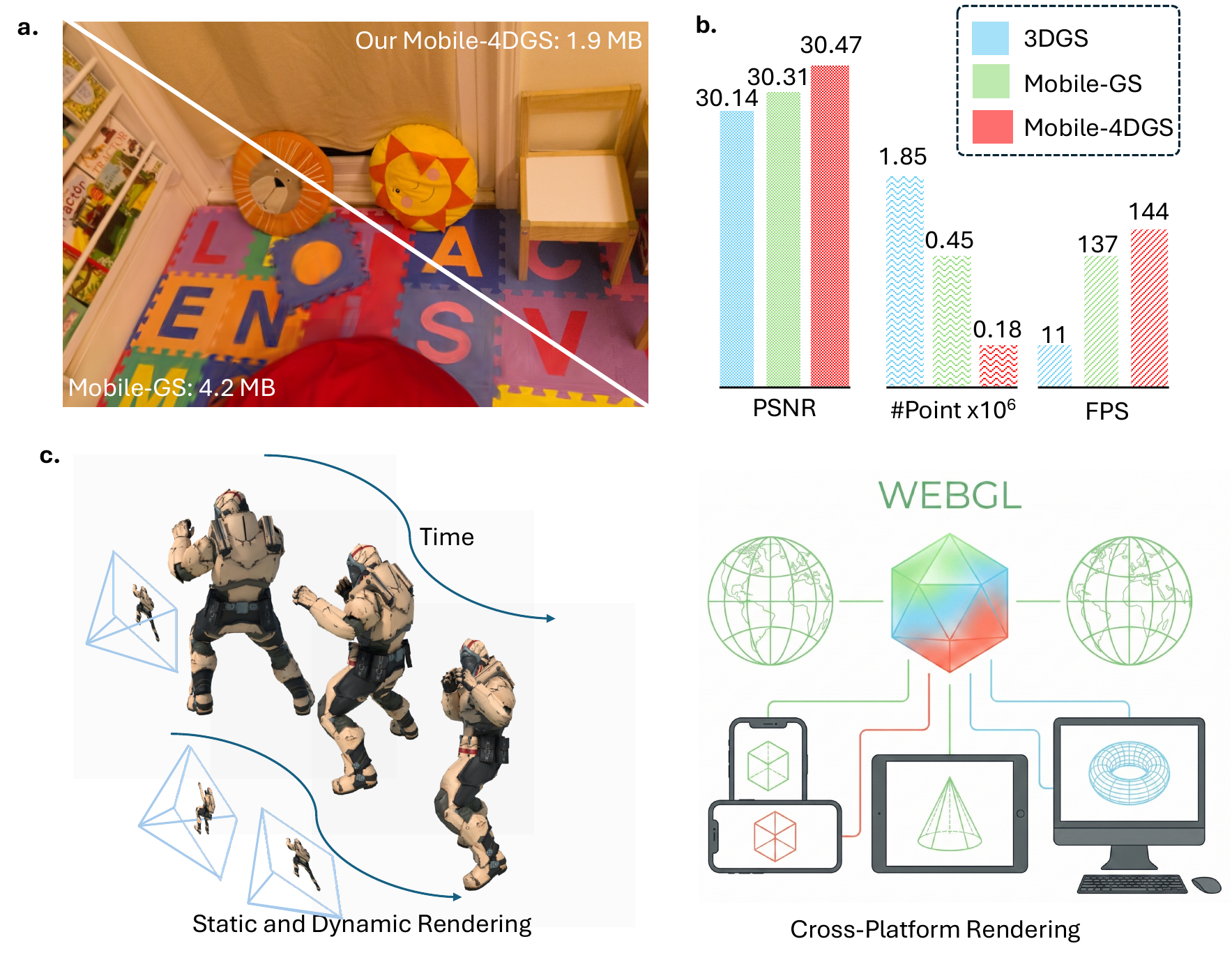}
    
    \caption{ \textbf{(a)(b) Mobile-4DGS} achieves rendering quality comparable to both 3DGS~\cite{kerbl2023gaussiansplatting} and Mobile-GS~\cite{mobilegs}, while reducing the number of Gaussian primitives and achieving significantly higher FPS on a mobile device with a Snapdragon 8 Gen 3 GPU. 
    \textbf{(c)} The proposed Mobile-4DGS utilizes WebGL to enable seamless cross-platform rendering and support both static and dynamic novel view rendering. }
    \vspace{-3mm}
    \label{fig:teaser}

\end{figure}

\section{Introduction}
Recent advancements in 3D scene reconstruction~\cite{pami1,pami2,pami4,pami6,yifan2019differentiable, npc} and novel view synthesis~\cite{nerfstudio,mildenhall2021nerf,chen2023mobilenerf,pami5,pami3,pami7} have been driven by 3D Gaussian Splatting (3DGS)~\cite{kerbl2023gaussiansplatting}. By representing scenes as millions of explicit and differentiable 3D Gaussians, 3DGS delivers high-fidelity, real-time rendering on desktop hardware. This capability has enabled diverse applications across autonomous driving~\cite{yan2024streetgs, du2024dreamcar, du20243drealcar}, navigation~\cite{du2020learning, du2023object, du2021vtnet}, and virtual reality~\cite{vrgs, long2023wonder3d, liu2023one, shen2023anything}. Despite these successes, efficiently deploying static and dynamic Gaussian Splatting on resource-constrained mobile devices remains a critical bottleneck.

Mobile-GS~\cite{mobilegs} is the first to deploy 3DGS on mobile devices and achieves real-time rendering. It identifies the sorting process in the original volume rendering as causing slow rendering for mobiles.
Therefore, it proposes an order-independent rendering pipeline to replace the traditional volume rendering so as to achieve fast rendering on mobiles. However, removing the sorting process causes Gaussian rendering in an arbitrary order, causing transparency artifacts in regions with overlapping geometry.
Although Flux-GS~\cite{du2026monte} maintains the original sorting process and achieves faster rendering, it cannot support dynamic rendering~\cite{4dgs} in our daily 4D space. The two lines of work therefore leave the same tension unresolved: giving up the sort buys throughput at the price of correct compositing, whereas keeping it means paying for a depth order that a 4D scene has to rebuild at every playback frame, together with the upload of the resulting index buffer.

In our experiments, we identified four primary bottlenecks that hinder the efficient deployment of Gaussian Splatting (3DGS) on mobile devices:
(1) \textbf{Third-Order Spherical Harmonics Overhead:} Prior methods~\cite{hanson2024speedy, lee2024compact, zhang2025gaussianspa, liu2025maskgaussian, hou2024sort} commonly rely on third-order Spherical Harmonics (SH) to model complex, view-dependent radiance. However, third-order SH requires 48 floating-point parameters per Gaussian primitive. Scaled across millions of Gaussians, this creates prohibitive storage overhead and heavy memory bandwidth consumption during rasterization.
(2) \textbf{Redundant Primitive Densification:} Lightweight 3DGS variants, such as Mobile-GS~\cite{mobilegs} and MEGS2~\cite{chen2025MEGS2}, typically adopt standard single-view gradient-based densification~\cite{kerbl2023gaussiansplatting}. This approach aggressively spawns excess primitives, leading to overfitting and inflated point counts. The resulting primitive redundancy severely impacts training efficiency, storage footprint, and inference latency on the resource-constrained edge devices.
(3) \textbf{MLP-based Dynamic Representation:}
Previous works~\cite{4dgs,sc-gs,yang2024deformable} leverage neural networks to model temporal variation, introducing non-trivial per-frame deformation inference and limiting rendering efficiency on resource-constrained devices.
(4) \textbf{Per-Frame Depth Ordering under Motion:}
Primitive motion continuously changes the relative depth of Gaussians, invalidating previously cached depth orders. As a result, the renderer must repeatedly re-project primitives, recompute their ordering, and update the draw list during playback, introducing additional runtime and data-transfer overhead.

\begin{figure*}[t]
    \centering
    \includegraphics[width=0.99\linewidth]{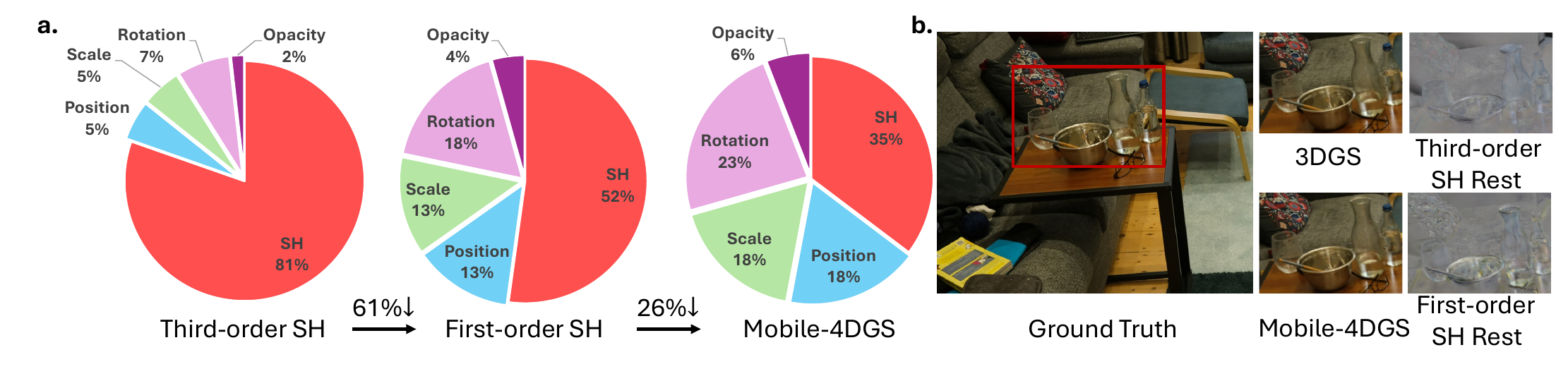}
    
    \caption{\textbf{Gaussian parameter distribution and Spherical Harmonic fidelity analysis. (a)} Per Gaussian memory footprint across 3DGS variants. Mobile-4DGS achieves significant compression (61\% and 26\% reductions) by optimizing Spherical Harmonics (SH) coefficients and decoupling SH into the base and view-independent components. 
    \textbf{(b)} Qualitative comparison demonstrates that Mobile-4DGS with only first-order SH can render high-fidelity high-frequency details comparable to 3DGS.}
        \vspace{-3mm}

    \label{fig:teaser_sh}
  
\end{figure*}

In this work, we present Mobile-4DGS, a mobile real-time Gaussian Splatting method designed to deliver high-fidelity rendering with significantly fewer parameters. 
As illustrated in Fig.~\ref{fig:teaser}, our proposed Mobile-4DGS achieves rendering quality comparable to existing baselines while utilizing fewer Gaussian primitives, thereby outperforming Mobile-GS~\cite{mobilegs} in inference speed. 
Our framework is built upon four technical innovations:
(1) \textbf{Unified Mobile Static-Dynamic Representation}: We extend the static Mobile-GS/Flux-GS foundations to a compact continuous-time representation in which each dynamic Gaussian requires only eight additional temporal scalars. Static scenes are represented as the zero-motion special case, allowing both modes to share the same appearance, compression, density-control, and rendering pipeline.
(2) \textbf{Monte Carlo Specular Energy Aggregator:} As shown in Fig.~\ref{fig:teaser_sh}, our approach is motivated by the observation that substantial radiance information can be compressed into a lower-order subspace without compromising view-dependent specular features.
 We propose a Monte Carlo Specular Energy Aggregator that aggregates high-order specular energy into a compact representation, effectively preserving the complex lighting characteristics.
Furthermore, we decouple the SH representation into a view-independent base component and directional view-dependent components, allowing for a substantial reduction in SH memory overhead compared to traditional third-order baselines without the need for expensive offline distillation or pre-training.
To recover the fine detail lost during this compression, we introduce an Attribute-Conditioned SH Enhancement module. This module utilizes a lightweight Multi-Layer Perceptron (MLP) to model contextual offsets based on intrinsic Gaussian properties. More importantly, these offsets are statically baked into the explicit Gaussian parameters prior to inference, ensuring that this enhanced view-dependent modeling does not introduce additional computational overhead during inference.
(3) \textbf{Multi-view Alpha-based Densification and Pruning strategy: }  We further address the inefficiency inherent in the standard single-view gradient-based densification strategy, which often lacks multi-view structure consistency. We propose a Multi-view Alpha-based Densification and Pruning strategy that employs stratified camera sampling and alpha-weighted error accumulation across multiple viewpoints. 
With multi-view guidance, this strategy ensures a more compact representation by accurately pruning redundant primitives while focusing densification on regions with high multi-view reconstruction error.
(4) \textbf{Depth-Order Certificate for 4D Playback:} The learned partition is exactly binary, so the depth order of the static subset depends on the camera alone and is cached for the whole clip, while a closed-form bound on the travel of the animated subset yields a per-frame depth-order certificate (DOC) that decides whether the committed draw list is still the sorted one. A frame that passes the certificate skips re-projection, sorting, merging and the index upload.

Extensive experiments demonstrate that Mobile-4DGS offers a robust and scalable solution for mobile real-time rendering. By achieving powerful first-order SH representation while maintaining competitive visual quality, our framework paves the way for the deployment of high-fidelity 3DGS on mobile devices. The same framework also keeps 4D playback affordable on the device: the exact certificate leaves the rendered image untouched on every frame it reuses, and the budget branch confines the reordering it admits to primitives whose footprints already overlap.

\noindent\textbf{Relationship to Prior Conference Versions.}
This paper substantially extends our previous Mobile-GS~\cite{mobilegs} and Flux-GS~\cite{du2026monte} papers. Mobile-GS established a mobile-oriented Gaussian rendering pipeline together with a distillation-based appearance representation, whereas Flux-GS investigated Monte Carlo aggregation for compact static Gaussian Splatting. The present journal version unifies these foundations under a single deployment-oriented framework and extends them to dynamic scenes through explicit second-order motion, learnable temporal support, temporally stratified optimization, and quantized temporal attributes. It further incorporates a shared cross-platform rendering pipeline for both static and dynamic scenes. Importantly, static rendering is treated as a zero-motion special case of the same explicit representation. We extend the sorting pipeline of Flux-GS~\cite{du2026monte}: because the learned gate is exactly the partition the renderer consumes, the per-camera cache of the static depth order and the depth-order certificate that licenses reuse of the committed draw list follow from the representation itself rather than from a separate engineering effort. 
We summarize our contributions as follows:

\begin{itemize}

\item  We extend the framework to 4D by incorporating non-linear temporal modeling and dynamic opacity lifespans, enabling real-time, high-quality 4D scene rendering on mobile platforms.

    \item   We propose a Monte Carlo aggregation strategy that decouples SH components to compress high-order view-dependent energy into a lower-order subspace, paired with an Attribute-Conditioned SH Enhancement module whose offsets are statically baked in before deployment to eliminate runtime inference overhead.

\item  We propose a multi-view densification and pruning strategy using stratified camera sampling and alpha-weighted error accumulation, ensuring structural consistency across views while reducing redundant Gaussians.

\item We propose a depth-order certificate (DOC) for mobile 4D playback, which caches the depth order of the static subset per camera and decides with a closed-form travel bound whether a committed draw list is still valid, so that certified frames skip re-projection, sorting, merging, and the index upload.

\end{itemize}

\section{Related work}

\noindent\textbf{Neural Radiance Fields.}
Recent advances in novel view synthesis have been heavily driven by 3D neural representations. Initial approaches associated neural features with underlying geometric structures such as volumes~\cite{sitzmann2019deepvoxels}, textures~\cite{chen2020neural}, or point clouds~\cite{kopanas2021point}. A major paradigm shift occurred with Neural Radiance Fields (NeRF)~\cite{mildenhall2021nerf}, which utilize multi-layer perceptrons (MLPs) to directly query 3D density and color, enabling photorealistic rendering from 2D images without explicit proxy geometry.
Subsequent research has focused on enhancing the rendering quality and computational efficiency of NeRF through optimized sampling~\cite{instant-ngp}, tensor decomposition~\cite{TensoRF}, compact representations~\cite{zhang2020nerf++}, and mobile-friendly pipelines~\cite{chen2023mobilenerf}. Despite these improvements, the heavy computational and memory footprints of NeRF-based models remain prohibitive for real-time deployment on resource-constrained edge devices.

\noindent\textbf{3D Gaussian Splatting.}
3D Gaussian Splatting (3DGS)~\cite{kerbl2023gaussiansplatting} has emerged as a powerful alternative, representing scenes via anisotropic 3D Gaussians and utilizing a tile-based rasterizer for real-time rendering. To scale this approach, recent works focus on accelerating training and managing memory.
Scaffold-GS~\cite{lu2024scaffold} introduces a hierarchical scaffold structure to reduce the number of Gaussians for high-quality rendering, while maintaining visual fidelity.
Octree-GS~\cite{ren2024octree} proposes a Level-of-Detail (LOD) structure for 3D Gaussian supporting level-of-detail decomposition.
3D-HGS~\cite{li2025halfgaussian} introduces a 3D-Half Gaussian as a novel reconstruction kernel to better learn the surface normal in 3D scenes.
3DGS-MCMC~\cite{mcmc} formulates the training dynamics as a Markov Chain Monte Carlo process, while Revisiting Densification~\cite{rota2024revising} proposes a loss-driven densification scheme. 
3DGS-LM~\cite{hollein20253dgs} utilizes a tailored Levenberg–Marquardt method~\cite{ranganathan2004levenberg} to replace the traditional Adam optimizer~\cite{zhang2018improved}, while Taming 3DGS~\cite{mallick2024taming} enforces a strict user-defined Gaussian budget during optimization.
MVGS~\cite{du2024mvgs} is the first to propose multi-view learning to enhance the multi-view constraint of 3DGS in the optimization stage, which significantly improves the holistic rendering performance.
Furthermore, techniques like StopThePop~\cite{radl2024stopthepop} and FlashGS~\cite{feng2025flashgs} tackle tile-intersection bottlenecks and depth-sorting challenges to mitigate popping artifacts and reduce computation complexity.

\noindent\textbf{Gaussian Pruning.}
Due to the inherent redundancy in dense 3D Gaussian representations~\cite{niemeyer2025radsplat, 3dgsdr, 2dgs}, pruning has become a key strategy to boost rendering efficiency~\cite{fan2024lightgaussian, mini-splatting, pixelgs, pcgs}. Common pruning strategies rely on importance scores derived from accumulated transmittance~\cite{girish2024eagles, ren2026fastgs}, Hessian-based sensitivity~\cite{hanson2025pup}, or gradient-opacity combinations~\cite{liu2024compgs, ali2025elmgs}.
While Mobile-GS~\cite{mobilegs} compresses the scene using opacity- and scale-based pruning, it struggles to preserve fine-grained details. Furthermore, standard single-view gradient-based densification tends to produce an excessive number of Gaussians. To address this, our proposed Mobile-4DGS introduces a multi-view alpha-based densification and pruning strategy, significantly minimizing the primitive budget while maintaining high rendering quality.

\noindent\textbf{Gaussian Compression.}
Early compression methods utilized vector quantization (VQ)~\cite{lee2024compact, papantonakis2024reducing, xie2024mesongs, omg} and entropy coding~\cite{chen2024hac, chen2025hac++} to lower storage requirements. While highly effective, large codebooks in VQ-based methods lead to long training times.  Although residual vector quantization (R-VQ)~\cite{liu2024compgs} mitigates this overhead, it incurs additional index storage costs. Recent work has shifted toward structured representations, combining anchor-based structures~\cite{lu2024scaffold, wang2024contextgs, liu2024compgs} or tensor factorization~\cite{f-3dgs} with grid hierarchies~\cite{chen2024hac, ren2024octree}. Contextual modeling~\cite{chen2024hac, chen2025hac++, wang2024contextgs, li2026gscodec} has also been introduced to maximize rate-distortion performance, though at the cost of high rendering latency due to multiple MLP passes.
Alternatively, neural field representations capture local continuity among neighboring Gaussians~\cite{lee2024compact, shin2025locality}. For instance, Mobile-GS~\cite{mobilegs} distills spherical harmonics (SH) from a high-order teacher, but this process is computationally expensive. In contrast, our Mobile-4DGS introduces a Monte Carlo Specular Energy Aggregator that directly compresses third-order SH energy into a compact, low-order representation, bypassing costly distillation while preserving high-fidelity view-dependent details.

\noindent\textbf{4D Gaussian Splatting.}
To reconstruct dynamic scenes, recent methods extend static representations into the temporal domain by modeling time-dependent geometry and appearance~\cite{4dgs}. Early attempts directly integrated time as a fourth coordinate or used dynamic multi-layer perceptrons (MLPs) to deform standard 3D Gaussians over time~\cite{gan2023v4d}. 
However, relying heavily on neural networks for deformation often introduces a significant computational bottleneck, hindering real-time rendering speeds during inference.
To address this, subsequent architectures have introduced explicit temporal structures, such as hexplane~\cite{cao2023hexplane} or 4D grid decompositions~\cite{gan2023v4d}, to separate spatial and temporal features efficiently. Other approaches leverage physical constraints or motion trajectories to track individual Gaussian primitives across frames, preserving fine-grained temporal consistency~\cite{4d-roto}. Despite these innovations, existing 4D architectures typically suffer from slow rendering speed on mobiles due to the volume of temporal tracking coefficients~\cite{zhang2025mega,omg4}.
In contrast, our approach optimizes temporal feature aggregation to maintain high-fidelity dynamic rendering without compromising efficiency on mobile devices.

\section{Preliminary}

\noindent\textbf{Gaussian Splatting:}
3D Gaussian Splatting (3DGS)~\cite{kerbl2023gaussiansplatting} is an explicit Gaussian-based rendering method that represents a scene using a collection of anisotropic 3D Gaussians, enabling high-quality novel view synthesis. 
Each 3D Gaussian \( \mathcal{G}_i \) contains 3D geometry and appearance parameters, as \( (\mathbf{\mu}_i, \Sigma_i, o_i,c_i) \). 
For the geometry representation, \( \mathbf{\mu}_i \in \mathbb{R}^3 \) defines the 3D position of the 3D Gaussian in its world coordinates, while the covariance matrix \( \Sigma_i \in \mathbb{R}^{3 \times 3} \), which is always positive semi-definite and symmetric, characterizes the spatial extent and orientation of the anisotropic Gaussian ellipsoid. 
In terms of geometric modeling, \( \Sigma_i \) is decomposed into a scale component \( s_i \) and a rotation component \( r_i \), where \( s_i \) controls the size of the Gaussian and \( r_i \) specifies its orientation.
As for appearance, each Gaussian leverages a color $ c_i(\omega) $ and an opacity value \( o_i \in [0,1] \), where Gaussian color \( c_i \) is represented using the spherical harmonic coefficients \(c_i \).

To render an image, all 3D Gaussians are projected onto the image space using a standard perspective camera model. 
After that, a depth-sorting procedure is then applied to ensure that $\mathcal{N} $ Gaussians are rendered in a near-to-far order. Following this ordering, the final color of a pixel is computed via the alpha blending:
\begin{equation}
    \mathbf{C} = \sum_{i=1}^{\mathcal{N}}  c_i(\omega) \alpha_i T_i,
\end{equation}
\begin{equation}
    \quad T_i = \prod_{j=1}^{i-1} (1 - \alpha_j),  
\end{equation}
\begin{equation}
  \alpha_i(\mathbf{x})
=
o_i
\exp\left(
-\frac{1}{2}
(\mathbf{x}-\boldsymbol{\mu}^{2D}_i)^\top
(\boldsymbol{\Sigma}^{2D}_i)^{-1}
(\mathbf{x}-\boldsymbol{\mu}^{2D}_i)
\right).
\end{equation}
where \( T_i \) denotes the accumulated transmittance before the \( i \)-th Gaussian.
\( \mathcal{N} \) is the total number of Gaussians contributing to the pixel.
Overall, Gaussian Splatting provides a differentiable, efficient, and compact representation for complex scenes, making it well suited for real-time rendering.

\noindent\textbf{Spherical Harmonics:}
Spherical harmonics provide an orthonormal basis for square-integrable functions defined on the unit sphere $\mathbb{S}^2$. Any sufficiently smooth directional function $c(\boldsymbol{\omega}) : \mathbb{S}^2 \rightarrow \mathbb{R}^3$ can be expanded as
\begin{equation}
c(\boldsymbol{\omega}) = \sum_{\ell=0}^L \sum_{m=-\ell}^{\ell} c_{\ell m} Y_{\ell m}(\boldsymbol{\omega}),
\end{equation}
where $Y_{\ell m}$ denotes the real spherical harmonics basis functions, and $c_{\ell m}$ are the corresponding SH coefficients.
In practice, the expansion is truncated to a finite maximum degree $L$, yielding $(L+1)^2$ coefficients. In modern neural rendering systems, low-order SH (e.g., $L \leq 2$) are often preferred due to their compactness and numerical stability, while higher-order SH are sometimes used during training to capture high-frequency view-dependent effects.

\section{Methodology}
Vanilla 3D Gaussian Splatting (3DGS) leverages explicit 3D Gaussians with volume rendering to represent a whole scene or object. Each 3D Gaussian \( \mathcal{G}_i \) is defined by a set of geometry and appearance parameters: \( (\mathbf{\mu}_i, s_i, r_i, o_i, c_i) \).
Specifically, $\mathbf{\mu}_i$, $s_i$, and $r_i$ represent spatial position, scale, and rotation, while $o_i$ and $c_i$ denote opacity and spherical harmonic coefficients for appearance modeling.
To capture complex and view-dependent effects, standard 3DGS methods typically utilize third-order spherical harmonics (SH).  
While effective for modeling specular and intricate lighting, this approach incurs significant storage costs, as each Gaussian primitive requires 48 SH coefficients ($16 \text{ coefficients} \times 3 \text{ channels}$). 
As scene complexity grows into the millions of primitives, the resulting storage footprint and memory bandwidth requirements become prohibitive for mobiles.

In this work, we propose Mobile-4DGS, which constrains the Gaussian appearance modeling to first-order spherical harmonics (requiring only 4 coefficients per channel). To ensure rendering fidelity, we propose a Monte Carlo Specular Energy Aggregator method to compress high-order radiance energy into a compact and lower-order subspace without pretraining or distillation.
By bridging the gap between high-fidelity radiance modeling and hardware-efficient representation, Mobile-4DGS enables high-quality and efficient novel view synthesis performance on edge devices.

\begin{figure*}[t!]
    \centering
    \includegraphics[width=0.99\linewidth]{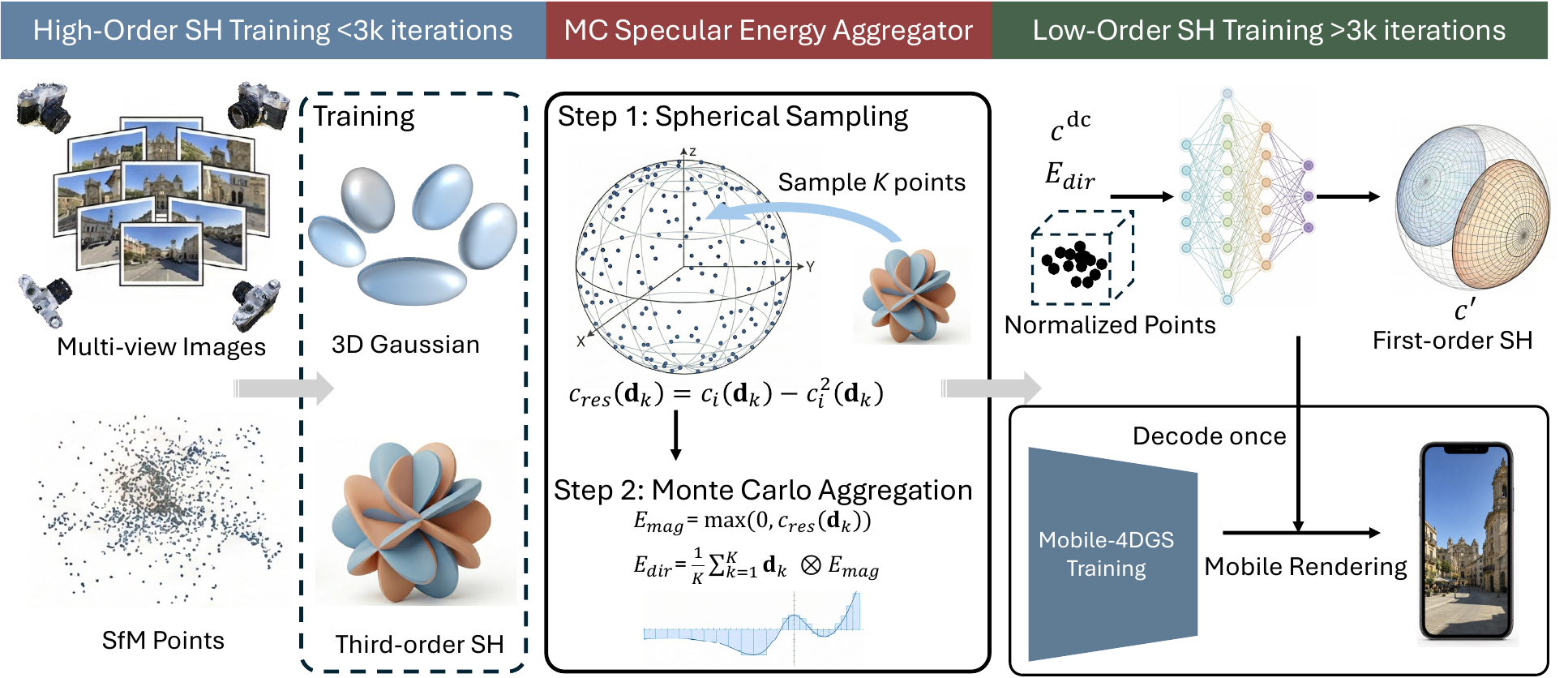}
    
    \caption{\textbf{Overview of the Mobile-4DGS framework.} Our method optimizes third-order SH for the initial 3k iterations, then transitions via Monte Carlo Specular Energy Aggregator for high-frequency representation. With the first-order direction moments inherited from the original third-order SH, we leverage a neural network to aggregate these latents into first-order SH $c'$ for rendering.  
    During inference, the model requires only a one-time decoding step to obtain the first-order SH coefficients, significantly reducing storage while introducing no per-frame decoding overhead.
   }
       \vspace{-3mm}

    \label{fig:method}
   
\end{figure*}

\subsection{Monte Carlo Specular Energy Aggregator}
\noindent\textcolor{teal}{$\blacktriangleright$~\textbf{``Insight 1":} Extracting Monte Carlo directional moments of specular residuals into pre-baked first-order parameters preserves complex lighting with zero runtime inference overhead.}

\noindent High-order SH polynomials in 3DGS are inherently inefficient for representing sharp, highly localized specular lobes due to their dense and global nature. Fitting a sparse specular peak requires numerous high-order SH merely to cancel out non-specular ringing artifacts. We observe that view-dependent energy is highly sparse in the angular domain, yet its directional phase is critical for accurate reconstruction. Rather than aggregating absolute photometric residuals, which destroys this vital angular gradient, we propose to extract the first-order directional moments of the specular residual via Monte Carlo projection.  It projects the extracted directional energy into the 1st-order SH space. Consequently, this insight allows us to discard the redundant high-order SH while explicitly preserving both the magnitude and spatial direction of the view-dependent signal.

Consider an RGB-valued directional function $c_i(\boldsymbol{\omega}) = \sum_{\ell=0}^{L_o}\sum_{m=-\ell}^{\ell} c_{\ell m} Y_{\ell m}(\boldsymbol{\omega})$ defined over the sphere $\mathbb{S}^2$. 
This function is initially represented by spherical harmonics of an original degree $L_o$, with a coefficient tensor: $c \in \mathbb{R}^{N \times (L_o+1)^2 \times 3}$ where $N$ represents the number of Gaussian primitives. Our objective is to compress this high-fidelity representation into a lower-order SH space of degree $L_t$ (where $L_t \leq L_o$), resulting in a reduced coefficient set: $c' \in \mathbb{R}^{N \times (L_t+1)^2 \times 3}$.
For each primitive $i$, we formulate this as a least-squares optimization problem:
\begin{equation}
c'^* = \arg\min_{c'} \int_{\mathbb{S}^2} \left| \left|\ c_i(\omega)  - \sum_{\ell=0}^{L_t} \sum_{m=-\ell}^{\ell} c'_{\ell m} Y_{\ell m} (\boldsymbol{\omega}) \right|\right|_2^2 d\boldsymbol{\omega},
\end{equation}
where $Y_{\ell m}(\boldsymbol{\omega})$ denotes a spherical harmonic basis function. 
To obtain a low-order representation, the optimal projection coefficients $c'_{n,k}$ that minimize the reconstruction error are given analytically by the inner product:
\begin{equation}
c'_{i,\ell m} = \langle c_i(\boldsymbol{\omega}) , Y_{\ell m} \rangle=\int_{\mathbb{S}^2}
c_i(\boldsymbol{\omega}) Y_{\ell m}(\boldsymbol{\omega}) 
d\boldsymbol{\omega},
\label{eq:project}
\end{equation}
where it 
defines the optimal orthogonal projection within the strict low-order SH subspace.  However, this projection cannot retain information that lies exclusively in the discarded higher-order bands. Our Monte Carlo module is therefore not introduced to approximate the same projection. Instead, it estimates compact directional statistics from the discarded residual and uses them to reparameterize the first-order coefficients before deployment.
To retain these high-frequency characteristics without incurring the prohibitive high-order SH, we aim to compress the discarded specular information into a compact latent representation.
The perceptual impact of specular is fundamentally characterized by its photometric energy magnitude and dominant direction. We instead propose to sample the high-frequency residual on a uniform sphere then extract its first-order directional moments.

\noindent\textbf{Uniform Spherical Sampling:}
To generate a set of directional vectors $\mathbf{D} = \{\mathbf{d}_k\}_{k=1}^K$, we uniformly sample $K$ points on the unit sphere $\mathbb{S}^2$ using spherical coordinates. For each sample $k$, we draw two independent random variables $\theta_k$ and $\phi_k$ respectively representing the colatitude and longitude:
\begin{equation}
   \quad \phi_k \sim \mathcal{U}(0, 2\pi), \quad  \theta_k = \arccos(1 - 2\xi_k), \quad \xi_k \sim \mathcal{U}(0, 1),
\end{equation}
where $\mathcal{U}$ denotes the uniform sampling.
These spherical coordinates are subsequently transformed into the Cartesian domain to obtain the unit direction vectors 
$
\mathbf{d}_k = \begin{bmatrix}
\sin \theta_k \cos \phi_k, ~
\sin \theta_k \sin \phi_k, ~
\cos \theta_k
\end{bmatrix}
$
where each $\mathbf{d}_k$ represents a unit vector such that $\|\mathbf{d}_k\|_2 = 1$. This collection of vectors $\mathbf{D} \in \mathbb{R}^{K \times 3}$ serves as the set of incident directions for subsequent spherical harmonic compression.

\noindent\textbf{Energy Aggregation:}
Subsequently, we evaluate the photometric residual between the third-order SH radiance $c_{i}(\mathbf{d}_k)$ and the second-order SH radiance $c_{i}^{2}(\mathbf{d}_k)$ via $c_{res}(\mathbf{d}_k) = c_{i}(\mathbf{d}_k) - c_{i}^{2}(\mathbf{d}_k)$. By aggregating these residuals, we formulate a compact, high-frequency dynamic energy and direction representation:
\begin{equation}
E_{mag} =  \max(0, c_{res}(\mathbf{d}_k)), 
\end{equation}
\begin{equation}
E_{dir} = \frac{1}{K} \sum_{k=1}^{K} \mathbf{d}_k \otimes E_{mag},
\label{eq:sh_hq}
\end{equation}
where $E_{mag}$ and $E_{dir}$ effectively encapsulate the overall energy and direction of the view-dependent specular for each Gaussian primitive, compressing the bulky high-order parameters into a highly lightweight low-dimensional descriptor.
To strictly minimize memory overhead, we only store these compact latents $E_{dir}$ and $c^{dc}$, where $c^{dc}$ represents the base color SH, as the 0-th SH. We then employ a neural network to learn a non-linear mapping from this aggregated high-frequency latent space to the first-order SH parameters, conditioned on the normalized Gaussian position:
\begin{equation}
    c' = \Psi(c^{dc},E_{dir}, f(\hat{\mu})),
\end{equation}
\begin{equation}
f(\hat{\mu}) = 
\begin{cases} 
\hat{\mu} & \text{if } \|\hat{\mu}\|_2 \leq 1, \\
\left( 2 - \frac{1}{\|\hat{\mu}\|_2} \right) \frac{\hat{\mu}}{\|\hat{\mu}\|_2} & \text{if } \|\hat{\mu}\|_2 > 1,
\end{cases}
\end{equation}
where $\Psi$ is a Multi-Layer Perceptron (MLP), and the result $c' \in \mathbb{R}^{N \times (L_t+1)^2 \times 3}$ represents the first-order SH coefficient ($L_t=1$). 
We normalize the Gaussian position $\mu$ to reduce the impact of spatial outliers for better convergence.
By performing this mapping, Mobile-4DGS can dynamically adjust the Gaussian geometry and low-order appearance simultaneously, ensuring that the restricted SH capacity is allocated to the most visually salient scene features. 
In particular, our approach eliminates the need for expensive distillation while achieving significant compression in SH storage memory overhead compared to third-order baselines. More importantly, the first-order SH coefficients are decoded only once before inference, introducing no per-frame decoding overhead.

\noindent\textbf{Gaussian Attribute Awareness:}
Although the above process preserves the majority of radiance energy in an $L^2$ sense, projecting higher-order bands inevitably degrades overall color fidelity and scene structure.
We observe that this discarded residual energy is highly correlated with intrinsic Gaussian attributes, such as spatial position and anisotropic scale.
This motivates us to design a lightweight geometry-conditioned residual module to contextually refine the base representation and preserve the efficiency of first-order SH.
Specifically, we introduce an Attribute-Conditioned SH Enhancement module parameterized by a lightweight MLP, denoted as $\Phi$, to predict a residual offset $\Delta c$ for the first-order SH coefficients $c'$. The network aggregates the geometric and photometric properties by concatenating the normalized first-order SH $\hat{c}'$, opacity $\alpha$, normalized scales $\hat{s}$, spatial position $\mu$, and rotation quaternion $r$. The enhanced SH coefficients $c^{out}$ are formulated as:
\begin{equation}
    \Delta c_i = \Phi([\mathbf{\mu}_i, \hat{s}_i, r_i, o_i, \hat{c}'_i ]), \quad c^{out}_i = c'_i + \Delta c_i,
\end{equation}
where the final linear layer of $\Phi$ is strictly zero-initialized to ensure optimization stability, allowing the model to naturally fall back to the standard 3DGS representation at the beginning of training and smoothly learn the residuals. Crucially, because $\Phi$ relies strictly on intrinsic Gaussian attributes and is entirely independent of the camera viewing direction, the residual $\Delta c$ only needs to be decoded once for subsequent inference. Prior to rendering, the SH offset is pre-computed and statically baked into the explicit Gaussian parameters. This design guarantees that our contextual modeling facilitates higher performance while introducing absolutely zero computational overhead during inference.

\noindent\textbf{Training Curriculum:} 
Initially, we optimize 3D Gaussians utilizing full third-order SH coefficients for the first 3k iterations to establish a high-frequency representation. Subsequently, we aggregate these third-order coefficients down to a first-order SH representation via our proposed Monte Carlo Specular Energy Aggregator. After this mapping step, we initialize the Attribute-Conditioned SH Enhancement module to continually refine the low-order representation.
Note that the compression from third-order to first-order is executed only once. After that, we only maintain and update the lightweight first-order parameters.

\begin{figure*}[t]
    \centering
    \includegraphics[width=0.99\linewidth]{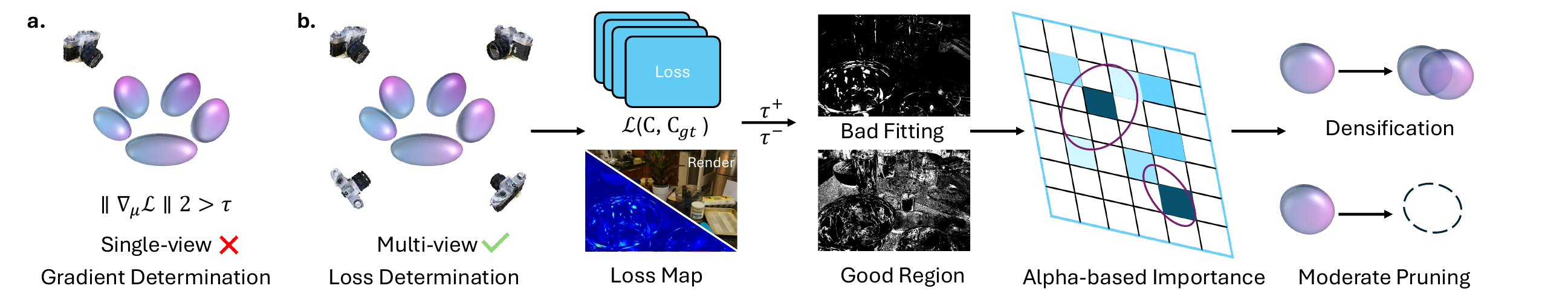}
    
    \caption{\textbf{Illustration of our proposed Multi-view Alpha-based Densification and Pruning strategy.} \textbf{(a)} Traditional 3DGS leverages the single-view Gaussian position gradient determinist. \textbf{(b)} We propose to employ a multi-view loss-driven mechanism, coupled with alpha-based Gaussian evaluation to densify more Gaussians in the bad-reconstructed regions and moderately prune Gaussians for well-reconstructed areas. }
        \vspace{-3mm}

    \label{fig:densify}
  
\end{figure*}

\subsection{Multi-view Alpha-based Densification and Pruning}
\noindent\textcolor{teal}{$\blacktriangleright$~\textbf{``Insight 2":} Stratified multi-view alpha-error accumulation maintains global structural consistency while pruning redundant primitives to prevent excessive Gaussian point inflation.}

\noindent  To achieve a compact 3D Gaussian structure, we found that an accurate densification and pruning strategy is crucial. We observed previous methods~\cite{kerbl2023gaussiansplatting, mini-splatting} using gradient-based or importance-based strategies in the single-view criterion to determine Gaussians' densification and pruning. However, the single-view criteria lack a holistic understanding of the scene, as they account for geometry and appearance from only a single view. Therefore, we propose a multi-view alpha-based densification and pruning strategy in Fig.~\ref{fig:densify} to constrain the number of Gaussians based on multi-view evaluation so that ensure multi-view consistency.

\noindent\textbf{Stratified Camera Sampling:}
To ensure comprehensive geometric coverage and alleviate view redundancy within the training camera set, we adopt a \textit{Stratified Camera Sampling} strategy. We partition the 3D scene into angular bins defined with respect to a focal point, enabling the selection of representative views that maximize view variation and angular dispersion.
Prior to binning, we estimate the center of the scene, denoted as $\mathbf{x}^*$. Given $N_{cam}$ cameras with the centers $\{\mathbf{t}_i\}_{i=1}^{N_{cam}}$ and viewing directions $\{\mathbf{v}_i\}_{i=1}^N$, we compute the point that minimizes the total squared distance to all camera rays. Let $P_i = I - \mathbf{v}_i \mathbf{v}_i^\top$ be the projection matrix onto the subspace orthogonal to the $i$-th ray, where $I$ is an identity matrix. The scene center $\mathbf{x}^*$ is obtained by solving the linear least-squares system via:
$
    \left( \sum_{i=1}^{N_{cam}} P_i \right) \mathbf{x}^* 
= 
\sum_{i=1}^{N_{cam}} P_i \mathbf{t}_i,
$
where each camera position is transformed into spherical coordinates $(\theta, \phi, \rho)$. $\theta = \mathrm{arctan2}(r_y, r_x)$, $\phi = \arcsin(r_z / \|\mathbf{r}_i\|)$, and $\rho = \|\mathbf{r}_i\|$, with $\mathbf{r}_i = \mathbf{t}_i - \mathbf{x}^*$. The angular space is discretized into $M_\theta$ azimuth and $M_\phi$ elevation bins. Cameras are grouped according to their bin indices, and one representative is uniformly sampled from each non-empty bin. Finally, to satisfy a computational budget $K$, the selected candidates are randomly shuffled and truncated, yielding a camera subset $\mathcal{C}$ with broad and uniform angular coverage.

\noindent\textbf{Photometric Loss Guidance:}
To effectively optimize the Gaussian Splatting representation, we employ a multi-view consistency-driven strategy. This strategy identifies regions of high and low reconstruction error across multiple viewpoints and uses the accumulated contribution of each Gaussian to guide geometry and appearance refinement. For each viewpoint in a curated camera set $\mathcal{C}$, we render the current Gaussian primitives and compute a pixel-wise photometric loss. Following standard 3DGS, we utilize the loss with a hybrid objective combining $L_1$ distance and Structural Similarity (SSIM) through:
\begin{equation}
\mathcal{L} = (1 - \lambda) \mathcal{L}_1(\mathbf{C}, \mathbf{C}_\text{gt}) + \lambda \mathcal{L}_{\text{DSSIM}}(\mathbf{C}, \mathbf{C}_\text{gt}),
\label{eq_loss}
\end{equation}
where $\mathbf{C}$ and $\mathbf{C}_\text{gt}$ represent the rendered image and Ground Truth. $\lambda$ controls the weight for $\mathcal{L}_1$ and $\mathcal{L}_{\text{DSSIM}}$.
Therefore, we can derive a binary Metric Map $M \in \{0, 1\}^{H \times W}$ to identify areas of poor and high-quality reconstruction through:
\begin{equation}
        M_{u,v}^{+} = \mathbb{I}\left( \mathcal{L}_{u,v}> \tau^{+} \right), \quad 
    M_{u,v}^{-}= \mathbb{I}\left( \mathcal{L}_{u,v} < \tau^{-}  \right),
\end{equation}
where $\mathbb{I}(\cdot)$ is the indicator function and $\tau$ is the error threshold.
A pixel $(u, v)$ is flagged ($M_{u,v}^+ = 1$) if its error exceeds a predefined threshold $\tau^+$, marking it as a region of poor reconstruction. On the contrary, $M_{u,v}^-=1$ means that the region is well reconstructed, needed to prune redundant Gaussians.

\noindent\textbf{Multi-view Alpha-Weighted Error Accumulation:}
To map 2D image-space errors back to 3D Gaussian primitives, we customize the CUDA kernel for the rendering pipeline. For each Gaussian $i$, we compute the importance score $S_i^+$ and pruning score $S_i^-$ by summing its alpha contributions $\alpha_{i}$ across all flagged pixels in the metric map by:
\begin{equation}
    S_i^+ = \sum_{c=0}^\mathcal{C} \sum_{(u,v) \in \Omega } \alpha_{i} \cdot M_{uv}^{c,+}, \quad
        S_i^- = \sum_{c=0}^\mathcal{C} \sum_{(u,v) \in \Omega } \mathcal{L}_{uv} \cdot \alpha_{i} \cdot M_{uv}^{c,-},
\end{equation}
where  $\alpha_i = o_i \exp\left(-\frac{1}{2}\Delta x_i^T \Sigma_i^{-1} \Delta x_i\right)$ represents Gaussian alpha, indicating the Gaussian contribution to the pixel color. $\Omega \subset \mathbb{Z}^2$ denote the discrete image pixel domain. We aggregate these local metrics across the sampled views to compute these two primary scores. This multi-view aggregation effectively differentiates Gaussians that are important or redundant, allowing for more accurate geometry modification. By averaging across the sampling views, our Mobile-4DGS avoids over-fitting to single-view occlusions or transient artifacts.
By using the alpha-based metric rather than the conventional gradient metric~\cite{kerbl2023gaussiansplatting}, we ensure that only the Gaussians with high visibility and contributing to the error are modified.

\noindent\textbf{Accurate Densification:}
The vanilla 3DGS employs a gradient-based strategy to identify underfitting Gaussians. We define the vanilla clone mask $\mathcal{M}_{clone}^{base}$ and split mask $\mathcal{M}_{split}^{base}$ based on standard positional gradient threshold $\|\nabla_{\mu_i} \mathcal{L}\|_2 > \tau$. 
However, relying solely on single-view positional gradients frequently results in excessive and redundant Gaussian densification.
To mitigate this issue, we integrate our computed importance score into the traditional 3DGS gradient-based densification strategy. Our multi-view consistency-guided masks are formed by intersecting these base criteria with an importance quantile filter:
\begin{equation}
    \mathcal{M}_{clone} = \mathcal{M}_{clone}^{base} \land \left( S^+  > Q_\tau^{+}(S^+)   \right),
\end{equation}
\begin{equation}
    \mathcal{M}_{split} = \mathcal{M}_{split}^{base} \land \left( S^+  >  Q_\tau^{+}(S^+) \right),
\end{equation}
where $ Q_\tau^{+}$ denotes the quantile of the importance score distribution. This ensures that Gaussians are only cloned or split if they consistently contribute to high-error regions across multiple views. Consequently, these refined masks strictly govern the cloning and splitting operations, boosting Gaussians' fitting capability while tightly controlling the primitive count.

\noindent\textbf{Moderate Pruning:}
Standard 3DGS defines a threshold $o_{min}$ to eliminate Gaussian points with low opacity, leading to aggressive pruning and performance slump. To achieve moderate pruning, we introduce the pruning score $S^-$ based on multi-view and loss-driven strategy. A larger $S^{-}$ indicates the Gaussian that impacts more to low-error pixel regions, effectively flagging it as an artifact or redundant point. Our final pruning mask $\mathcal{M}_{prune}$ identifies Gaussians that fall below the opacity threshold $o_{min}$ and exceed the pruning score quantile $Q_\tau^{-}$ via:
\begin{equation}
    \mathcal{M}_{prune} = (o_i < o_{min}) \land \left( S^- >  Q_\tau^{-}(S^-) \right),
\end{equation}
where we utilize $\mathcal{M}_{prune}$ to cull non-contributing or artifact-inducing Gaussians while preserving essential scene geometry. In this way, we achieve moderate pruning while maintaining overall reconstruction fidelity.

\begin{figure*}[t!]
    \centering
        \includegraphics[width=0.89\linewidth]{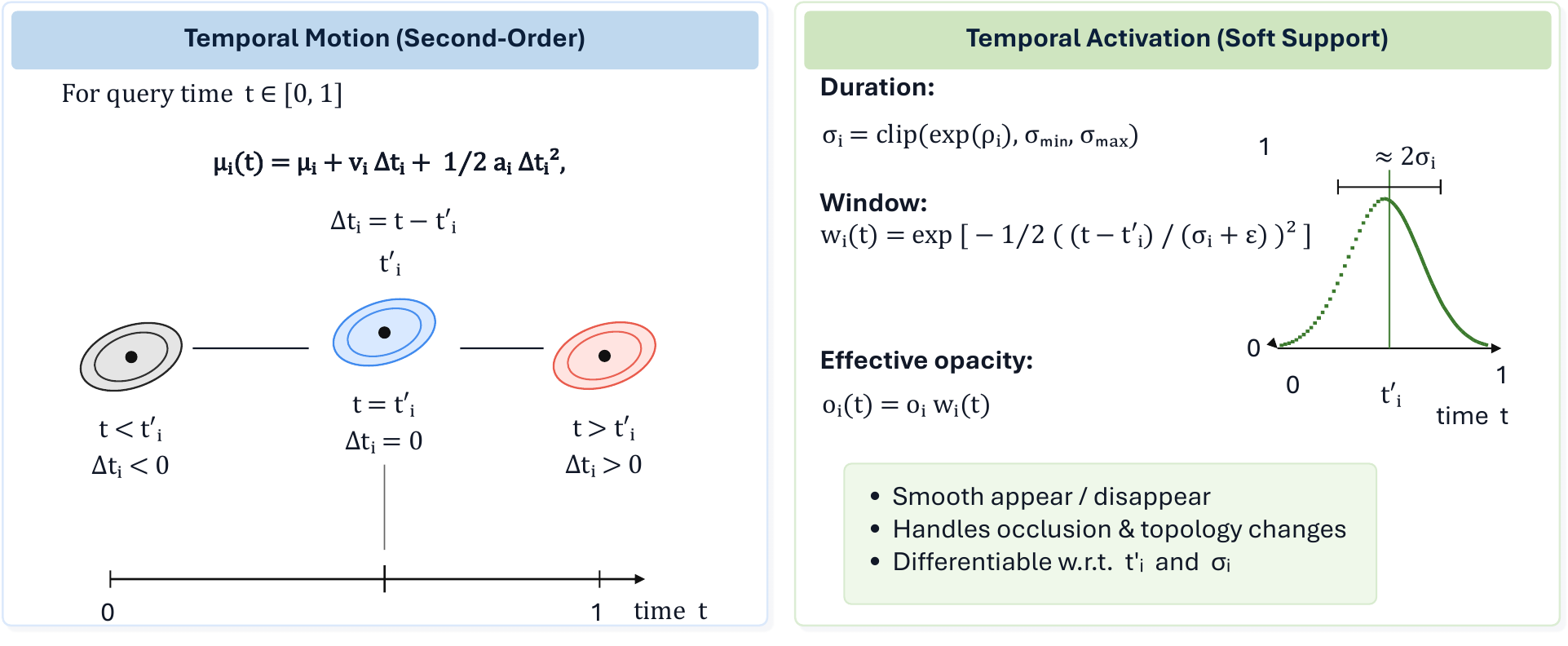}

    \caption{\textbf{Overview of the temporal modeling in Mobile-4DGS.}
Each Gaussian is equipped with an explicit second-order motion state $(\mathbf{v}_i,\mathbf{a}_i,t'_i,\rho_i)$ and a learned binary static--dynamic gate $g_i$. Given a query time $t\in[0,1]$, its center is updated from the canonical state by Eq.~\eqref{eq:temporal_motion}, so that the displacement of a primitive with $g_i=0$ is identically zero rather than merely small. A Gaussian temporal window, parameterized by the learnable duration $\sigma_i$, modulates its effective opacity through Eq.~\eqref{eq:temporal_opacity}, enabling smooth appearance and disappearance while handling occlusions and topology changes. The entire formulation is differentiable with respect to the motion and temporal parameters, and the committed partition exposes an exactly time-invariant subset that can be cached and compressed separately.}
\vspace{-3mm}
    \label{fig:method_4d}

\end{figure*}

\subsection{Dynamic Modeling}

\noindent\textcolor{teal}{$\blacktriangleright$~\textbf{``Insight 3":}
A learned binary static--dynamic partition allows each Gaussian to model
continuous-time motion with simple vector operations, while the static subset
remains exactly unchanged over time and can therefore be cached for mobile
rendering.}

\noindent
As illustrated in Fig.~\ref{fig:method_4d}, we extend each static Gaussian with
a compact set of temporal parameters. We normalize the timestamp of each input
frame to $t\in[0,1]$. In addition to its canonical center
$\boldsymbol{\mu}_i\in\mathbb{R}^{3}$, covariance
$\boldsymbol{\Sigma}_i$, base opacity $o_i$, and appearance coefficients
$\mathbf{c}_i$, each Gaussian $\mathcal{G}_i$ stores a velocity
$\mathbf{v}_i\in\mathbb{R}^{3}$, an acceleration
$\mathbf{a}_i\in\mathbb{R}^{3}$, a canonical time $t'_i\in[0,1]$, and a
log-duration $\rho_i\in\mathbb{R}$. This adds only eight scalars per Gaussian.
More importantly, all temporal states are explicit, so no deformation network
needs to be evaluated during rendering. This avoids the per-frame neural
inference required by implicit deformation-based methods.

\noindent\textbf{Gated Second-Order Trajectory:}
We model the position of each Gaussian at time $t$ using a second-order motion
model around its canonical time:
\begin{equation}
  \boldsymbol{\mu}_i(t)
    = \boldsymbol{\mu}_i
    + g_i\left(\mathbf{v}_i\,\Delta t_i
    + \frac{1}{2}\mathbf{a}_i\,\Delta t_i^2\right),
    \qquad
    \Delta t_i = t-t'_i ,
    \label{eq:temporal_motion}
\end{equation}
where the velocity term models approximately constant motion, while the
acceleration term captures changes in speed or direction. The binary gate
$g_i\in\{0,1\}$ determines whether the Gaussian is static or dynamic. When
$g_i=0$, the entire displacement is exactly zero, and the Gaussian remains at
its canonical position for all timestamps. When $g_i=1$, its position follows
the learned trajectory. This gives a continuous trajectory over the whole
sequence instead of predicting a separate state for every frame. We keep
$\boldsymbol{\Sigma}_i$ and $\mathbf{c}_i$ constant over time to further reduce
temporal storage and computation.

\noindent\textbf{Soft Temporal Support:}
A dynamic Gaussian does not necessarily need to contribute to every frame.
We therefore assign each Gaussian a positive temporal duration
\begin{equation}
    \sigma_i = \operatorname{clip}\left(\exp(\rho_i),\,\sigma_{\min},\,\sigma_{\max}\right),
    \label{eq:temporal_duration}
\end{equation}
and define its temporal weight as
\begin{equation}
    w_i(t) =
    \exp\left[-\frac{1}{2}
    \left(\frac{t-t'_i}{\sigma_i+\varepsilon}\right)^2\right],
    \label{eq:temporal_window}
\end{equation}
where $\varepsilon$ prevents division by zero. We then define the effective
opacity as
\begin{equation}
    o_i(t) = \big[(1-g_i) + g_i\,w_i(t)\big]\,o_i ,
    \label{eq:temporal_opacity}
\end{equation}
so that static Gaussians always keep their original opacity, while dynamic
Gaussians smoothly appear and disappear around their canonical time $t'_i$.
During rendering, we simply replace $o_i$ with $o_i(t)$ in the standard alpha
compositing process. This temporal window helps model visibility changes,
occlusions, and topology changes while remaining differentiable with respect
to $\mathbf{v}_i$, $\mathbf{a}_i$, $t'_i$, and $\rho_i$.

\noindent\textbf{Learned Static--Dynamic Partition:}
Most Gaussians in a dynamic scene usually belong to static regions such as the
background, floor, and stationary objects. Modeling temporal variation for
these Gaussians is unnecessary. On the other hand, incorrectly treating a
moving Gaussian as static can leave a frozen copy of the moving object in the
scene. We therefore learn whether each Gaussian is static or dynamic instead
of assigning all Gaussians to the same category.

Specifically, we jointly optimize a binary gate $g_i$ and the motion parameters.
We use a hard-concrete relaxation with a straight-through estimator, which gives
a binary value in the forward pass while still allowing gradients to propagate
during training. Given $u_i\sim\mathcal{U}(0,1)$, we compute
\begin{equation}
\begin{aligned}
  s_i &= \operatorname{sigmoid}\!\left(
  \frac{\ell_i + \log\frac{u_i}{1-u_i}}{T}\right), \\
  \tilde{s}_i &= \operatorname{clamp}\!\big(
  s_i(\zeta-\gamma)+\gamma,\,0,\,1\big),
\end{aligned}
\label{eq:gate_relaxation}
\end{equation}
followed by
\begin{equation}
  g_i = \mathbb{I}\!\left[\tilde{s}_i \geq 0.5\right], \qquad
  \widehat{g}_i =
  \operatorname{sg}(g_i)
  - \operatorname{sg}(\tilde{s}_i)
  + \tilde{s}_i ,
  \label{eq:dynamic_gate}
\end{equation}
where $\ell_i$ is the learnable gate logit, $T$ is the temperature,
$(\gamma,\zeta)$ define the relaxed range $[-0.1,1.1]$,
$\mathbb{I}[\cdot]$ is the indicator function, and
$\operatorname{sg}(\cdot)$ denotes stop-gradient. Thus,
$\widehat{g}_i$ is exactly equal to the binary gate $g_i$ in the forward pass,
while its relaxed value provides gradients in the backward pass. During
training, $\widehat{g}_i$ is used in
Eqs.~\eqref{eq:temporal_motion} and \eqref{eq:temporal_opacity}.

We gradually anneal the temperature from
$T_{\mathrm{init}}=2.0$ to $T_{\mathrm{final}}=0.1$, allowing the gates to
start with a smooth relaxation and gradually become binary. The partition is
also preserved during densification. When a Gaussian is cloned or split, its
new copies inherit the gate logit, forced-dynamic flag, committed label, and
promotion evidence from the parent. Logistic noise is used only during the
first $5$k iterations. The gates are then evaluated deterministically, and the
static--dynamic partition is committed and frozen at iteration $25$k. The
renderer and compression stage therefore operate on a fixed binary partition.

\noindent\textbf{Motion Evidence and Promotion:}
The static--dynamic assignment is not determined by the sparsity prior alone.
We additionally use motion evidence to prevent moving Gaussians from being
incorrectly classified as static.
First, a Gaussian is forced to be dynamic when its initialized velocity
satisfies
$\|\mathbf{v}_i\|_2>\tau_m$. Once a Gaussian is forced or promoted to dynamic,
it cannot later be changed back to static. This avoids a common failure case in
which a moving Gaussian is turned static and leaves a frozen copy of the object
at its canonical position.
When velocity initialization is available, free Gaussians start with a strong
static prior and become dynamic only when supported by motion evidence. When no
velocity initialization is available, they instead start with a strong dynamic
prior, because an entirely static initialization would provide little learning
signal to the temporal parameters.

To recover Gaussians that are incorrectly classified as static, we maintain an
exponential moving average of their reconstruction evidence:
\begin{equation}
  e_i \leftarrow
  \eta\,e_i +
  (1-\eta)\,
  \operatorname{ReLU}\!\left(
  -\frac{\partial \mathcal{L}}{\partial \ell_i}\right),
  \label{eq:promotion_ema}
\end{equation}
where a negative gate gradient indicates that increasing the gate would reduce
the photometric reconstruction loss and therefore provides evidence that the
Gaussian should move. Before computing this evidence, we remove the gradient
contribution from the gate regularizers, so that the sparsity prior itself
cannot trigger a promotion. After a $2$k warmup, we check the static population
every $500$ iterations and promote Gaussians whose evidence is above the
$0.995$ quantile. To avoid unstable changes, at most $1\%$ of all Gaussians can
be promoted at each step.

\noindent\textbf{Gate Regularization:}
Without additional constraints, the optimization may assign every Gaussian to
the dynamic set. We therefore encourage a sparse dynamic partition using the
closed-form hard-concrete relaxation of the $\ell_0$ penalty:
\begin{equation}
\begin{aligned}
  \mathcal{L}_{\mathrm{sparse}}
  &= \frac{1}{|\mathcal{F}|}
  \sum_{i\in\mathcal{F}}
  \operatorname{sigmoid}\!\left(
  \ell_i - T\log\frac{-\gamma}{\zeta}\right), \\
  \mathcal{L}_{\mathrm{bin}}
  &= \frac{1}{|\mathcal{F}|}
  \sum_{i\in\mathcal{F}} p_i(1-p_i),
  \quad p_i = \operatorname{sigmoid}(\ell_i),
\end{aligned}
\label{eq:gate_sparsity}
\end{equation}
where $\mathcal{F}$ contains only free Gaussians, excluding those that have
already been forced or promoted to dynamic. The sparsity loss is gradually
introduced from iterations $5$k to $20$k, while the binary loss is activated
from iteration $22$k until the partition is committed at $25$k. The two
mechanisms play complementary roles: the sparsity loss encourages unnecessary
dynamic Gaussians to become static, while the promotion mechanism recovers
static Gaussians that still require motion to explain the observations.

\noindent\textbf{Temporal Regularization:}
Unconstrained motion can produce unstable trajectories, duplicated edges, and
long motion trails. We therefore regularize the temporal parameters of dynamic
Gaussians:
\begin{align}
  \mathcal{L}_{\mathrm{acc}}
  &= \frac{\sum_i g_i \|\mathbf{a}_i\|_2^2}
  {\sum_i g_i},
  \label{eq:acceleration_regularization}\\
  \mathcal{L}_{\mathrm{dur}}
  &= \frac{
  \sum_i g_i
  \operatorname{ReLU}\!\left(
  \sigma_i-\sigma_{\max}^{\mathrm{soft}}\right)^2}
  {\sum_i g_i},\\
  \mathcal{L}_{\mathrm{ext}}
  &= \frac{
  \sum_i g_i
  \operatorname{ReLU}\!\left(
  \|\mathbf{v}_i\|_2\,\sigma_i
  + \tfrac{1}{2}\|\mathbf{a}_i\|_2\,\sigma_i^{2}
  - d_{\max}\right)^2}
  {\sum_i g_i}.
  \label{eq:extent_regularization}
\end{align}
Because these losses are multiplied by $g_i$, they affect only dynamic
Gaussians and do not alter the canonical geometry of the static subset.
$\mathcal{L}_{\mathrm{acc}}$ discourages unnecessarily large acceleration,
while $\mathcal{L}_{\mathrm{dur}}$ prevents the temporal support from becoming
too broad. The motion-extent loss
$\mathcal{L}_{\mathrm{ext}}$ constrains the displacement reached within one
temporal standard deviation,
$\|\mathbf{v}_i\|_2\sigma_i+
\tfrac{1}{2}\|\mathbf{a}_i\|_2\sigma_i^2$,
which directly reduces motion trails without globally forcing all velocities
to be small.
Therefore, the complete objective is
\begin{equation}
\begin{aligned}
    \mathcal{L}_{4\mathrm{D}}
    &= \mathcal{L}
    + \lambda_{\mathrm{acc}}\mathcal{L}_{\mathrm{acc}}
    + \lambda_{\mathrm{dur}}\mathcal{L}_{\mathrm{dur}} \\
    &\quad
    + \lambda_{\mathrm{ext}}\mathcal{L}_{\mathrm{ext}}
    + \lambda_{\mathrm{sparse}}\mathcal{L}_{\mathrm{sparse}}
    + \lambda_{\mathrm{bin}}\mathcal{L}_{\mathrm{bin}}.
\end{aligned}
\label{eq:temporal_total_loss}
\end{equation}

\noindent\textbf{Staged Optimization:}
We jointly optimize the spatial, temporal, and appearance parameters from
multi-view videos. Different temporal parameters follow different learning-rate
schedules. The learning rates of acceleration, canonical time, and log-duration
are cosine-decayed to zero between iterations $25$k and $30$k to prevent the
temporal support from drifting late in training. The velocity instead follows
its own exponential decay toward a small final learning rate. Canonical times
are randomly initialized over the sequence, while the initial temporal
durations are set to a small fraction of the clip and constrained to
$[\sigma_{\min},\sigma_{\max}]$ throughout optimization.

After the binary partition is committed, we do not freeze the complete static
Gaussian. Instead, we freeze only its four time-varying parameters. Its
canonical position, scale, rotation, opacity, and appearance remain trainable
during the final refinement stage. This allows the static geometry and
appearance to continue improving even after its temporal state has been fixed.

\noindent\textbf{Temporal-Aware Density Control:}
For dynamic scenes, the multi-view densification and pruning procedure is also
extended to the temporal dimension. We divide the sequence into temporal bins,
sample one camera from each non-empty bin, and then subsample these candidates
to match the camera budget. In this way, short-lived Gaussians are observed
during density control even if they are visible in only a small part of the
sequence. This prevents such Gaussians from being incorrectly densified or
pruned simply because they are rarely observed.

\noindent\textbf{Compression of the Motion State:}
Each dynamic Gaussian stores the eight-dimensional temporal vector
$[\mathbf{v}_i,\mathbf{a}_i,t'_i,\rho_i]\in\mathbb{R}^{8}$.
We compress this vector using the same sub-vector codebook mechanism used for
the static attributes, while fitting the temporal codebooks only on dynamic
Gaussians. Because the partition is already binary, static Gaussians do not
consume temporal codebook capacity and instead share an index that decodes to
an exact zero-motion state.

Only compact codebook indices are stored for each Gaussian, and the resulting
index streams are entropy coded. We also store the committed binary partition
as a packed bitmask so that decoding exactly reproduces the static--dynamic
assignment used by the renderer. For backward compatibility, a checkpoint
without explicit gate bits is treated as fully dynamic. Temporal quantization
is performed after static quantization, followed by a separate refinement stage
for the temporal codebooks. This prevents static and temporal compression
errors from interfering with each other.

\noindent
At inference time, the temporal parameters are recovered through simple
codebook lookups. Equations~\eqref{eq:temporal_motion} and
\eqref{eq:temporal_opacity} then require only a quadratic motion evaluation and
a temporal-window evaluation for each dynamic Gaussian. No deformation network
is required, and the appearance and covariance parameters are decoded only
once rather than once per frame.

The binary static--dynamic partition also directly benefits rendering.
For $g_i=0$, the Gaussian position does not change with time, so its depth order
depends only on the camera and can be cached across the sequence. For $g_i=1$,
the explicit motion model allows us to bound how far a Gaussian can move along
the camera depth axis. As described in Sec.~\ref{sec:render}, this bound allows
the renderer to determine whether an existing draw list can safely be reused,
avoiding repeated depth sorting on many playback frames.

\subsection{Mobile Rendering of the Gated Representation}
\label{sec:render}

\noindent\textcolor{teal}{$\blacktriangleright$~\textbf{Insight~4.}
The binary partition separates static and dynamic Gaussians into two streams.
The static depth order can be cached for a fixed camera, while a motion bound
determines when the previously computed global draw order is still valid.
Frames that pass this test require no re-projection, sorting, merging, or index
upload.}

\noindent\textbf{Two-Stream Depth Order.}
We divide the Gaussians into a static set
$\mathcal{S}=\{i:g_i=0\}$ and a dynamic set
$\mathcal{D}=\{i:g_i=1\}$, with
$N_{\mathcal{S}}+N_{\mathcal{D}}=N$.
The two sets are sorted independently.
Let $\mathbf{n}\in\mathbb{R}^{3}$ be the depth row of the camera transform.
The view-space depth of Gaussian $i$ is
$z_i(t)=\mathbf{n}^{\!\top}\boldsymbol{\mu}_i(t)$.
For a fixed camera, both streams use the same depth range
$[z_{\min},z_{\max}]$. We quantize each depth into a 16-bit key:
\begin{equation}
  q_i(t)
  =
  \operatorname{clip}\!\left(
  \left\lfloor
  \kappa\bigl(z_i(t)-z_{\min}\bigr)
  \right\rfloor,
  0, B-1
  \right),
  \label{eq:depth_key}
\end{equation}
where
$\kappa=\frac{B-1}{z_{\max}-z_{\min}}$ and $B=2^{16}$.
A smaller key indicates a Gaussian that is closer to the camera.
For either stream
$\mathcal{A}\in\{\mathcal{S},\mathcal{D}\}$,
we use one stable counting-sort pass over the $B$ possible depth keys. Since the
input indices are initially ordered by row id, the stable sort also preserves
the row-id order when multiple Gaussians have the same key. Thus, each stream
is ordered exactly according to
\begin{equation}
  \pi_{\mathcal{A}}(t)
  =
  \operatorname{argsort}_{(q_i(t),\,i)}\ \mathcal{A}.
  \label{eq:stream_order}
\end{equation}
The two sorted streams are then merged with a standard two-pointer procedure
using the same pair $(q_i,i)$, producing
\begin{equation}
  \pi(t)
  =
  \operatorname{argsort}_{(q_i(t),\,i)}
  \bigl(\mathcal{S}\cup\mathcal{D}\bigr).
  \label{eq:draw_order}
\end{equation}
Therefore, independently sorting the static and dynamic streams and then
merging them gives exactly the same result as globally sorting all Gaussians.
Here, ``exact'' refers to the 16-bit quantized depth ordering of
Eq.~\eqref{eq:depth_key}.
The key advantage of this separation is that the static stream can be cached.
For $g_i=0$, Eq.~\eqref{eq:temporal_motion} gives
$\boldsymbol{\mu}_i(t)=\boldsymbol{\mu}_i$, so its depth key does not change
with time. For a fixed camera, the static stream therefore needs to be projected
and sorted only once.
If the camera depth mapping changes, i.e., either $\mathbf{n}$ or
$[z_{\min},z_{\max}]$ changes, both the static cache and the committed global
draw list are invalidated and recomputed. A tolerance-based camera-change test
may be used in the relaxed interactive mode, but it is not part of the exact
reuse guarantee described below.

\begin{algorithm}[t]
\caption{Adaptive depth-order reuse (worker thread, one playback frame at time $t$).}
\label{alg:reuse}
\begin{algorithmic}[1]
  \Require committed list $\pi_c$ and its time $t_c$, streams
  $\mathcal{S},\mathcal{D}$, bounds $V,A,t'_{\min},t'_{\max}$,
  mean radius $\bar{s}$, budget $r$
  \State $M \gets$ Eq.~\eqref{eq:movement_bound}
  \If{the camera depth mapping changed}
    \State re-project and sort $\mathcal{S}$; \Comment{refresh static cache}
    \State re-project and sort $\mathcal{D}$; merge; post $\pi$
    \State $\pi_c \gets \pi$; $t_c \gets t$
    \State recompute $g_{\min}$ if $r=0$
  \ElsIf{$r=0$ \textbf{and} $2M<g_{\min}$}
    \State acknowledge reuse; \Comment{certified for quantized ordering}
  \ElsIf{$r>0$ \textbf{and}
  $M\le\left\lceil\kappa r\bar{s}\|\mathbf{n}\|_2\right\rceil$}
    \State acknowledge reuse; \Comment{budget branch}
  \Else
    \State re-project and sort $\mathcal{D}$; merge; post $\pi$
    \State $\pi_c \gets \pi$; $t_c \gets t$
    \State recompute $g_{\min}$ if $r=0$
  \EndIf
\end{algorithmic}
\end{algorithm}

\begin{table*}[t]
  \caption{Cost of one playback frame. The $B=2^{16}$ counting-sort scratch
  space is reused across frames, while the static stream is re-projected and
  re-sorted only when the camera depth mapping changes.}
  \centering
  \begin{tabular}{lccc}
    \toprule
    frame & worker & main thread & payload \\
    \midrule
    committed, Eq.~\eqref{eq:draw_order}
      & $O(N+B)$: projection, sorts, merge
      & $O(N)$ index upload
      & $4N$ bytes \\
    reused, Eq.~\eqref{eq:budget_policy}
      & $O(1)$ certificate
      & ---
      & $O(1)$ \\
    \bottomrule
  \end{tabular}
  \label{tab:render_cost}
\end{table*}

\noindent\textbf{Depth Travel Bound.}
For a fixed camera, only Gaussians in the dynamic set $\mathcal{D}$ change
position over time. The static set $\mathcal{S}$ keeps the same depth keys.
Therefore, to determine whether the current draw order can be reused, we only
need to bound how far the dynamic Gaussians can move in depth.
From Eq.~\eqref{eq:temporal_motion}, the depth change of a dynamic Gaussian
between a committed time $t_c$ and a query time $t$ is
\begin{equation}
\begin{split}
  z_i(t)-z_i(t_c)
  &= (t-t_c)\,\mathbf{n}^{\top}\mathbf{v}_i \\
  &\quad + \frac{1}{2}(t-t_c)\,\mathbf{n}^{\top}\mathbf{a}_i
     \bigl(\Delta t_i(t)+\Delta t_i(t_c)\bigr),
\end{split}
\label{eq:depth_increment}
\end{equation}
where we use
$\Delta t_i(t)^2-\Delta t_i(t_c)^2
=(t-t_c)\bigl(\Delta t_i(t)+\Delta t_i(t_c)\bigr)$.
Computing this displacement for every dynamic Gaussian each frame would remove
the benefit of the certificate. We therefore precompute a few global bounds:
\begin{equation}
\begin{aligned}
  V &= \max_{i\in\mathcal{D}}\lVert\mathbf{v}_i\rVert_2,
  \qquad
  t'_{\min} = \min_{i\in\mathcal{D}}t'_i, \quad
  t'_{\max} = \max_{i\in\mathcal{D}}t'_i, \\
  A &= \max_{i\in\mathcal{D}}\lVert\mathbf{a}_i\rVert_2,
  \quad
  \tau(t) =
  \max\bigl(
  \lvert t-t'_{\min}\rvert,
  \lvert t-t'_{\max}\rvert
  \bigr).
\end{aligned}
\label{eq:tau_definition}
\end{equation}
Here, $V$ and $A$ are the maximum velocity and acceleration magnitudes among
all dynamic Gaussians, while $\tau(t)$ bounds their temporal offsets.
These quantities can be collected once from the decoded dynamic Gaussians and
do not depend on the camera or query time.
Using the Cauchy--Schwarz and triangle inequalities gives a common depth-motion
bound for every $i\in\mathcal{D}$:
\begin{equation}
\begin{split}
  \Delta_z(t)
  &= \lvert t-t_c\rvert\,\lVert\mathbf{n}\rVert_2 \\
  &\quad \times
  \Bigl(
  V + \frac{1}{2}A
  \bigl(\tau(t)+\tau(t_c)\bigr)
  \Bigr),
\end{split}
\label{eq:depth_travel}
\end{equation}
such that
\[
\lvert z_i(t)-z_i(t_c)\rvert\le\Delta_z(t).
\]
We then convert this depth bound into the quantized key space. Since
$\lvert\lfloor x\rfloor-\lfloor y\rfloor\rvert
\le\lceil\lvert x-y\rvert\rceil$
and clipping cannot increase the difference, every dynamic Gaussian satisfies
\begin{equation}
  \lvert q_i(t)-q_i(t_c)\rvert \le M(t),
  \qquad
  M(t)=\bigl\lceil\kappa\,\Delta_z(t)\bigr\rceil.
  \label{eq:movement_bound}
\end{equation}
Static Gaussians have zero key change. Importantly, $M(t)$ is computed only
from a few precomputed scalars and therefore has constant per-frame cost,
independent of the number of dynamic Gaussians.

\noindent\textbf{Exact Reuse.}
We now use the motion bound to determine whether the committed draw list can be
reused exactly. Let $\pi_c$ be the draw list computed at time $t_c$, and let
$\pi(t)$ be the draw list that would be obtained by sorting again at time $t$.
Only neighboring pairs involving at least one dynamic Gaussian can change their
relative order. We therefore define
\begin{equation}
\begin{aligned}
  \mathcal{K}_c
  &= \bigl\{\, k :
  \pi_c(k)\in\mathcal{D}
  \ \text{or}\
  \pi_c(k+1)\in\mathcal{D}
  \,\bigr\}, \\
  g_{\min}
  &= \min_{k\in\mathcal{K}_c}
  \bigl(
  q_{\pi_c(k+1)}(t_c)
  -q_{\pi_c(k)}(t_c)
  \bigr).
\end{aligned}
\label{eq:min_gap}
\end{equation}
Thus, $g_{\min}$ is the smallest depth-key gap between neighboring Gaussians
for which at least one Gaussian can move. Static--static pairs are ignored
because their relative order cannot change for a fixed camera. If two relevant
Gaussians already share the same key, then $g_{\min}=0$ and reuse is
conservatively rejected.
Because every dynamic key can move by at most $M(t)$, the committed order is
guaranteed to remain unchanged whenever
\begin{equation}
  2M(t) < g_{\min}.
  \label{eq:exact_reuse_condition}
\end{equation}
When this condition holds,
\[
  \pi_c=\pi(t),
\]
so reusing $\pi_c$ produces exactly the same 16-bit quantized draw order as
sorting the scene again.
This can be shown by contradiction. Assume that two adjacent Gaussians
$(u,v)$ in $\pi_c$ reverse their order at time $t$, such that
$q_u(t_c)<q_v(t_c)$ but $q_v(t)<q_u(t)$. At least one of them must be dynamic,
because the keys of a static--static pair do not change. Since each movable key
can change by at most $M(t)$,
\begin{equation}
\begin{aligned}
  q_v(t_c) \le q_v(t)+M(t)
  &< q_u(t)+M(t) \\
  &\le q_u(t_c)+2M(t).
\end{aligned}
\label{eq:inversion_chain}
\end{equation}
Therefore,
$q_v(t_c)-q_u(t_c)<2M(t)$,
which contradicts Eq.~\eqref{eq:exact_reuse_condition}. Hence no relevant
adjacent pair can reverse its order, and the committed draw list remains
identical to the newly sorted list. The strict inequality also prevents a
moving pair from collapsing to the same key, preserving the row-id tie-breaking
rule used by the renderer.

\noindent\textbf{Budget Reuse.}
The exact test can be conservative for dense scenes because it depends on the
smallest relevant depth-key gap. We therefore provide an optional relaxed mode
that measures the allowed depth motion relative to the mean Gaussian radius
$\bar{s}$. The complete reuse policy is
\begin{equation}
  \operatorname{reuse}(t)
  =
  \begin{cases}
    \left[\,2M(t)<g_{\min}\,\right],
    & r=0\quad\text{(certified)},\\[2pt]
    \left[\,M(t)\le
    \left\lceil\kappa\,r\,\bar{s}\,\|\mathbf{n}\|_2\right\rceil\,\right],
    & r>0\quad\text{(budget)}.
  \end{cases}
  \label{eq:budget_policy}
\end{equation}
Here, $\bar{s}$ is the mean Gaussian radius computed once from the decoded
scales, and $r$ specifies the tolerated depth motion in units of this radius.
The $r=0$ case gives the exact certificate of
Eq.~\eqref{eq:exact_reuse_condition}. For $r>0$, the test is a practical
relaxation and may allow small local ordering changes. We use $r=0.5$ mean
radii as the default setting in our mobile viewer.

\noindent
When a frame is reused, the worker directly keeps the committed draw list.
It performs no primitive re-projection, counting sort, or stream merge, and
returns only a constant-size acknowledgement. The main thread consequently
keeps the existing GPU index buffer and skips the index upload. Since a newly
committed draw list contains one 32-bit index per Gaussian, it requires $4N$
bytes of transfer, whereas a reused frame requires only $O(1)$ communication.
Table~\ref{tab:render_cost} summarizes the two cases.

\subsection{Implementation}


\noindent \textbf{Deployment on Mobiles}: 
To enable cross-platform real-time rendering, we introduce a WebGL-based rendering framework for our Mobile-4DGS. Unlike traditional CUDA-dependent pipelines, our approach leverages the ubiquitous WebGL API to achieve real-time rendering directly within browsers, without external dependencies. 
To address the non-commutative nature of alpha-compositing, we implement a decoupled sorting strategy. 
By offloading the depth-sorting of millions of Gaussians to an asynchronous WebWorker and using the CPU for the visibility order update, we can break through the 120 display FPS limit in the rendering loop. 
This framework significantly reduces the barrier for navigable 3D environments, providing a cross-platform solution for real-time rendering.

\begin{table*}[t!]
\centering

\caption{\textbf{Comprehensive evaluation on the mobile device with Snapdragon 8 Gen 3 GPU on the Mip-NeRF 360 dataset~\cite{mip-nerf360}.}
To facilitate a comprehensive analysis of Flux-GS, scenes are categorized into indoor and outdoor subsets.
\#G denotes the number of Gaussian primitives.
 3DGS* represents the quantized version through Huffman encoding for mobile rendering.
 Mobile-GS* denotes the version of Mobile-GS~\cite{mobilegs} without MLP in the inference stage.
 We compare with this version without MLP for fairness.
The \colorbox{green!30}{best} and \colorbox{lime!30}{second-best} results are highlighted.
}

\scriptsize
\def\arraystretch{1.2}
\resizebox{0.99\linewidth}{!}
{
    \begin{tabular}{ l  |ccccc|ccccc}
    \toprule
    Category& \multicolumn{5}{c|}{Indoor}  &  \multicolumn{5}{c}{Outdoor}  \\
    
    Method \& Metrics& PSNR$\uparrow$ & \#G $\times 10^6$$\downarrow$& Storage (MB)$\downarrow$&FPS$\uparrow$&Train (min)$\downarrow$& PSNR$\uparrow$ &  \#G $\times 10^6$$\downarrow$& Storage (MB)$\downarrow$&FPS$\uparrow$ &Train (min)$\downarrow$\\
      \midrule   \midrule 

    3DGS~\cite{kerbl2023gaussiansplatting} &30.41&1.45& 478& -&27&24.61& 3.14& 1361&  -&   36\\

    \midrule

    3DGS* &30.04&1.45& 46&  11&27&   24.39& 3.14& 85& 5&36\\
    
    

    Speedy-Splat~\cite{hanson2024speedy}  &30.11&0.37& 64& 21&14&  24.41& 0.56& 88& 14&  13  \\

    C3DGS~\cite{lee2024compact}  & 30.01& 0.75& 21 & 18&31&  24.38&  0.91&  34&  13&45\\

    LocoGS-S~\cite{shin2025locality}  &  30.08&0.82&  6.1&  13&46&  24.31& 1.41&  10&   19& 61\\

    Mobile-GS~\cite{mobilegs}    & \colorbox{lime!30}{30.37}& 0.38 & 3.5  &   131  & 86 & \colorbox{green!30}{24.51} &0.58 &5.5  &114 & 136 \\  

        Mobile-GS*   & 29.58& 0.38 & 3.4  &   \colorbox{lime!30}{142}& 64& 23.74&0.58 &5.4  &128& 114\\  

    Flux-GS    & 30.22 & \colorbox{green!30}{0.22} & \colorbox{green!30}{2.1 }& \colorbox{green!30}{147}  & \colorbox{green!30}{11} & \colorbox{lime!30}{24.45} &\colorbox{lime!30}{0.48}  &  \colorbox{lime!30}{4.6} &\colorbox{green!30}{132} & \colorbox{green!30}{11} \\    \midrule
    
    Mobile-4DGS (\textbf{Ours})    & \colorbox{green!30}{30.47} & \colorbox{lime!30}{0.27} & \colorbox{lime!30}{2.8 }& \colorbox{lime!30}{143}  & \colorbox{lime!30}{12} & 24.42 &\colorbox{green!30}{0.46}  &  \colorbox{green!30}{4.3} &\colorbox{lime!30}{130} & \colorbox{lime!30}{12} \\  \bottomrule
    \end{tabular}
}

\label{table:360}

\end{table*}





    
    



     

   
    

\begin{table*}[t!]
\centering

\caption{\textbf{Quantitative evaluation of state-of-the-art light-weight Gaussian-based methods on the real-world datasets.} We report performance on the Tank\&Temples~\cite{tanktemple} and Deep Blending~\cite{deepblending} datasets.  
 }
 
\scriptsize
\def\arraystretch{1.2}
\resizebox{0.99\linewidth}{!}
{
    \begin{tabular}{ l  |cccccc|cccccc}
    \toprule
    Dataset  & \multicolumn{6}{c|}{Tanks\&Temples}  &  \multicolumn{6}{c}{Deep Blending}  \\
    
    Method \& Metrics& PSNR$\uparrow$ & SSIM$\uparrow$ & LPIPS$\downarrow$  & Storage (MB) $\downarrow$&FPS$\uparrow$ &Train (min)$\downarrow$ & PSNR$\uparrow$ & SSIM$\uparrow$ & LPIPS$\downarrow$  & Storage (MB) $\downarrow$&FPS$\uparrow$ &Train (min)$\downarrow$ \\
      \midrule   \midrule 

    
    


    3DGS~\cite{kerbl2023gaussiansplatting} &23.14 & 0.841&0.183  & 358.7& -&34&29.41&0.903 & 0.243 & 697.3   &  -&   24\\    \midrule



    
    

    Speedy-Splat~\cite{hanson2024speedy}  &23.08 &0.821 &0.241 & 62.4  & 16&34&  29.11 & 0.864 &0.309    & 71.2  & 24&\colorbox{lime!30}{16}\\

    C3DGS~\cite{lee2024compact}  & \colorbox{lime!30}{23.32} & 0.831 &\colorbox{lime!30}{0.202} & 21.8  & 15&49& 29.73 & 0.900 & 0.258    &  24.7  &  16&35\\

    LocoGS-S~\cite{shin2025locality}  &  23.23 &\colorbox{green!30}{0.837} &\colorbox{green!30}{0.204} &  6.8  &  21&65&  29.76 & \colorbox{lime!30}{0.903} &\colorbox{lime!30}{0.251}    &  7.8  &   14& 47\\

    Mobile-GS~\cite{mobilegs}    & 23.09 & 0.831&0.208  & \colorbox{lime!30}{2.5}  &   124&  82 &\colorbox{green!30}{29.93}&\colorbox{green!30}{0.906} & \colorbox{green!30}{0.243} &4.6  &135&  75 \\  

    Mobile-GS*   & 22.41& 0.825&0.227& \colorbox{lime!30}{2.5}  &   129&  71&29.14&0.895& 0.281&\colorbox{lime!30}{4.5}  &146&  63\\  
        
    Flux-GS     & 23.27 &  0.827  &0.221  & \colorbox{green!30}{2.4}& \colorbox{green!30}{137}& \colorbox{green!30}{11} &29.73&0.898 & 0.275 &\colorbox{green!30}{1.9} &\colorbox{green!30}{158}& \colorbox{green!30}{10}\\   \midrule
    
 Mobile-4DGS (\textbf{Ours})    & \colorbox{green!30}{23.83} &  \colorbox{lime!30}{0.836}  &0.216  & 3.2& \colorbox{lime!30}{132}& \colorbox{lime!30}{12} &\colorbox{lime!30}{29.92}&0.902 & 0.272 &\colorbox{green!30}{2.2} &\colorbox{lime!30}{151}& \colorbox{green!30}{12}\\   \bottomrule

    \end{tabular}
}

\label{table:tant}

\end{table*}

\section{Experiments}

\subsection{Implementation Details}
We train our proposed Mobile-4DGS with 30k iterations, following the vanilla 3DGS, without pretraining or distillation. In the initial 3k iterations, we train with the third-order Spherical Harmonics. Then, we employ our proposed Monte Carlo Specular Energy Aggregator to obtain the first-order SH representation from the previous third-order SH.
For our proposed Monte Carlo Specular Energy Aggregator, we sample $K=2048$ uniform points on a unit sphere. We only exert it once to obtain low-order SH representation.
To obtain the first-order SH, we employ two MLPs initialized with one hidden layer (64 neurons) for the mapping.
As to our proposed Attribute-Conditioned SH Enhancement module, this module is parameterized by 4 layers of MLPs with ReLU activation. The hidden neurons for these MLPs are (128, 64, 32, 12), respectively. 
As for our proposed Multi-view Alpha-based Densification and Pruning strategy.
We sample 6 cameras for this strategy to make sure multi-view consistency.
For the loss thresholds, we set $\tau^+=0.1$ and $\tau^-=0.01$ to identify regions of well-reconstructed and poorly-reconstructed. 
As for the quantile for importance and pruning, we set $Q_\tau^+=0.6$ and  $Q_\tau^-=0.1$.
For the quantization, we leverage the same method as the previous Mobile-GS~\cite{mobilegs}.
We train our proposed Mobile-4DGS on an RTX 4090 GPU.

\noindent\textbf{Hardware Platform:}
We conduct all mobile-centric rendering experiments on a commercial smartphone equipped with the Qualcomm Snapdragon 8 Gen 3 GPU for static scenes and an iPhone 14/A15 for dynamic scenes.
To ensure fair and reproducible evaluation, we measure rendering performance utilizing an offscreen benchmarking protocol, which breaks the screen display 120 FPS limit and UI overhead. Specifically, frames are rendered to an offscreen frame buffer, and the average FPS is computed over multiple consecutive runs after a short warm-up period to mitigate thermal and initialization effects.
As for the desktop GPU, we use a single RTX 4090.


\begin{figure*}[t!]
    \centering
    \includegraphics[width=0.99\linewidth]{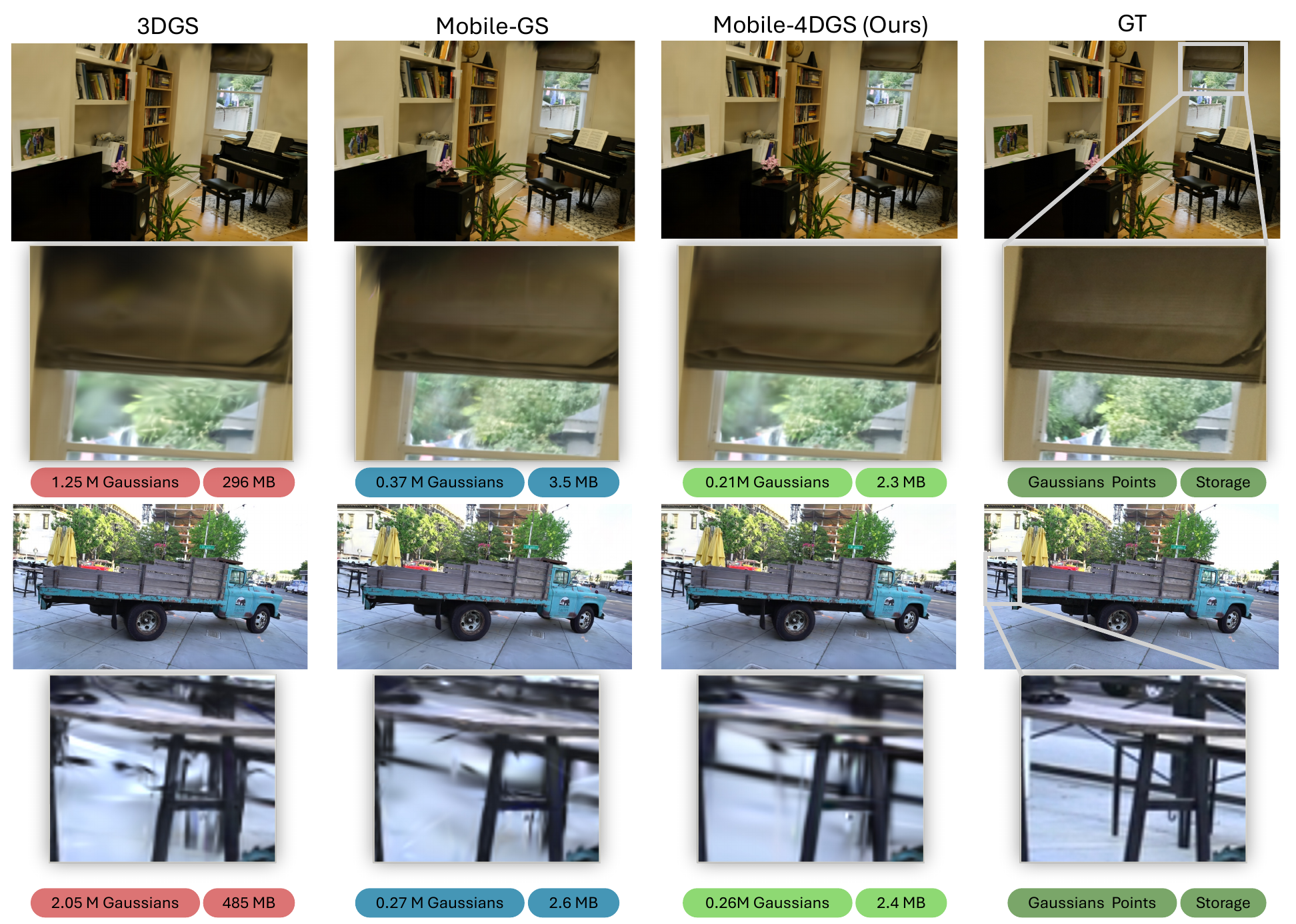}
        
    \caption{\textbf{Qualitative and efficiency comparison with previous state-of-the-art methods.}
We compare rendering quality, Gaussian number, and storage costs across 3DGS, Mobile-GS, and our Mobile-4DGS. 
Zoomed-in regions highlight details and structural consistency for clearer differentiation. 
Our method achieves comparable or superior visual fidelity while using significantly fewer Gaussians and substantially lower storage for real-time rendering on mobiles.
The number of Gaussians and total model size are reported below each result, demonstrating the improved efficiency–quality trade-off of our approach.
}
    \vspace{-3mm}

    \label{fig:nvs}
  
\end{figure*}

\begin{table}[t!]
    \centering
    \caption{\textbf{Quantitative performance on N3DV~\cite{n3dv} dataset. }  We report FPS results on an iPhone 14 smartphone with the A15 Bionic chip. The \colorbox{green!50}{best},  \colorbox{green!20}{second best}, and  \colorbox{lime!30}{third best} results are highlighted, respectively. }
    \resizebox{0.99\linewidth}{!}{
    \begin{tabular}{c|ccccccc}
    \toprule
    Method  & SSIM$\uparrow$  & SH $\downarrow$ & FPS$\uparrow$ &  \#Gaussian $\downarrow$ & Storage (MB)$\downarrow$ & Train (min) $\downarrow$  \\
    \midrule
    \midrule
    Real-Time4DGS~\cite{real-time4dgs} & \colorbox{green!50}{0.949} &\colorbox{lime!50}{3}&  \colorbox{lime!30}{3}  & \colorbox{lime!30}{2,770,350} & \colorbox{lime!30}{1701}  & \colorbox{green!20}{374} \\
 
    OMG4~\cite{omg4} & 	\colorbox{lime!30}{0.942} &\colorbox{lime!50}{3}& \colorbox{green!20}{12} &	\colorbox{green!50}{283,297} &	\colorbox{green!20}{5.8} & \colorbox{lime!30}{417} \\
    \midrule
     Mobile-4DGS (\textbf{Ours}) & 	\colorbox{green!20}{0.945}	 & \colorbox{green!50}{1}&\colorbox{green!50}{47} &	\colorbox{green!20}{285,707} &	\colorbox{green!50}{3.8} & \colorbox{green!50}{62} \\
    \bottomrule
    \end{tabular}
    }
    \label{tab:3dv}
    \vspace{-0.5cm} 
\end{table}
\begin{figure*}[t!]
    \centering
    \includegraphics[width=0.99\linewidth]{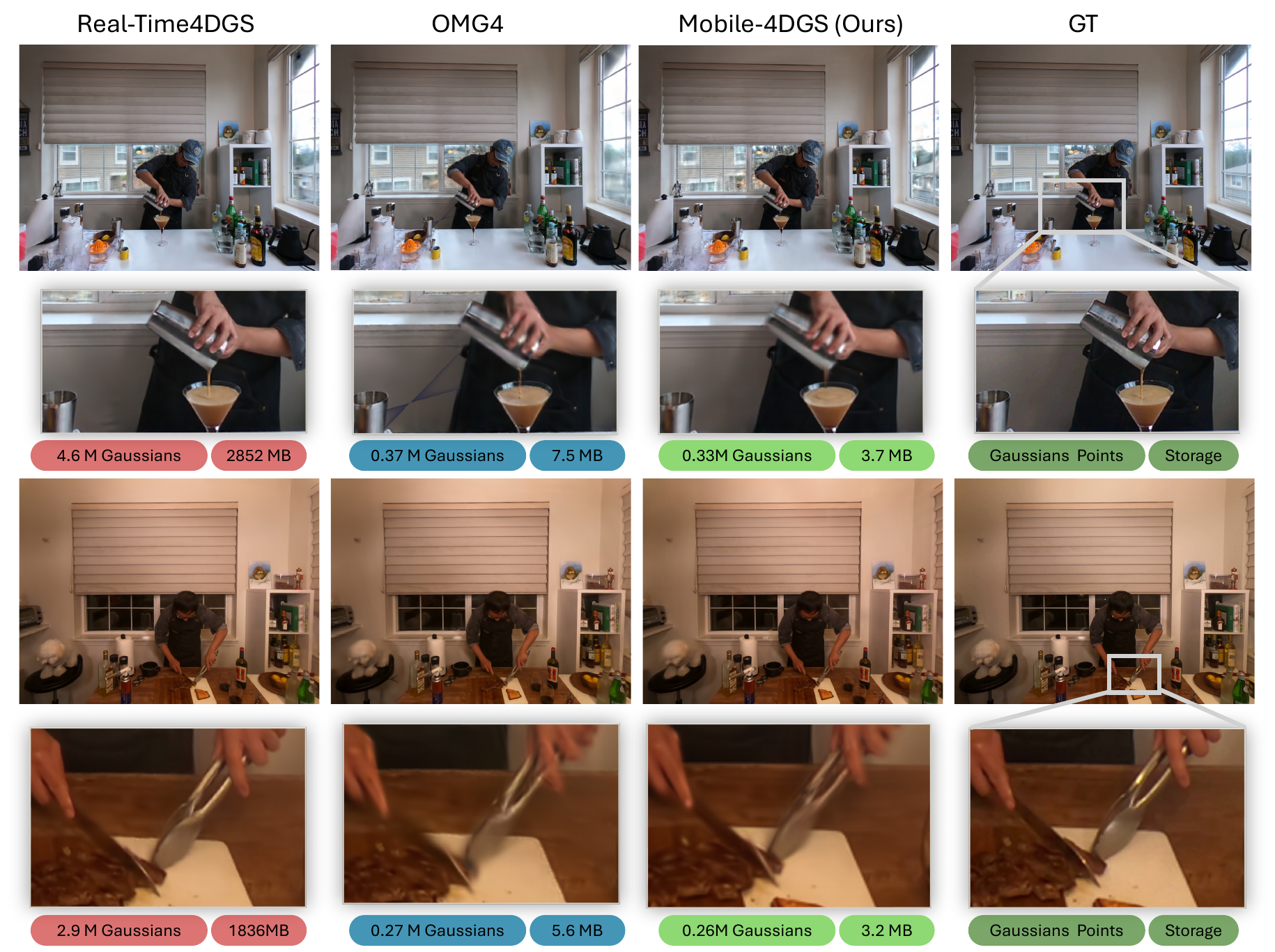}
        
    \caption{\textbf{Qualitative and efficiency comparison with previous state-of-the-art methods for dynamic scenes on the N3DV dataset~\cite{n3dv}.}
We compare rendering quality, Gaussian count, and storage overhead across Real-Time4DGS, OMG4, and Mobile-4DGS. 
}
    \label{fig:nvs_4d}
      \vspace{-3mm}

\end{figure*}

\subsection{Qualitative and Quantitative Results}

\noindent\textbf{Static Evaluation and Mobile Performance:}
As detailed in Tables~\ref{table:360} and~\ref{table:tant}, we evaluate Mobile-4DGS against state-of-the-art lightweight Gaussian Splatting variants, including 3DGS~\cite{kerbl2023gaussiansplatting}, Speedy-Splat~\cite{hanson2024speedy}, C3DGS~\cite{lee2024compact}, LocoGS~\cite{shin2025locality}, and Mobile-GS~\cite{mobilegs} across the Mip-NeRF 360~\cite{mip-nerf360}, Tanks and Temples~\cite{tanktemple}, and Deep Blending~\cite{deepblending} datasets. To ensure a fair evaluation of baseline rendering efficiency, we compare against a variant of Mobile-GS that omits the Multi-Layer Perceptron (MLP) during inference, as the MLP introduces significant computational overhead.
Notably, our proposed method achieves a significant breakthrough in the trade-off between rendering fidelity and computational efficiency.  The original 3DGS and its quantized counterpart 3DGS* incur substantial storage costs and exhibit slow rendering speed on the mobile platform. Mobile-4DGS maintains competitive image quality metrics while delivering fast FPS across all tested indoor and outdoor environments. Furthermore, by optimizing the Gaussian distribution and utilizing our proposed compressed first-order SH representation, our method achieves a low memory footprint, compressing entire scenes to very low storage costs. This minimal storage requirement makes it particularly well-suited for resource-constrained edge devices.
Beyond inference efficiency, a key advantage of Mobile-4DGS is its drastically reduced training time. 
Our model achieves convergence several times faster than prior state-of-the-art methods, including Speedy-Splat and Mobile-GS. This accelerated convergence is directly attributed to our proposed multi-view alpha-based densification and pruning strategy, which minimizes the number of Gaussian primitives during optimization.
Mobile-4DGS achieves state-of-the-art rendering efficiency, offering rapid training and minimal storage.

\noindent\textbf{Dynamic Assessment and Mobile Results:}
As shown in Table~\ref{tab:3dv}, Mobile-4DGS achieves a favorable balance between rendering quality and computational efficiency on the N3DV dataset~\cite{n3dv} when evaluated on an iPhone~14 equipped with the A15 Bionic chip. Although Real-Time4DGS provides the highest overall reconstruction fidelity, it requires substantially greater storage and computational resources, resulting in limited rendering performance on the mobile platform. In contrast, Mobile-4DGS delivers the highest rendering throughput, the smallest storage footprint, and the shortest training time while maintaining competitive SSIM performance. Compared with OMG4, our method significantly improves rendering and training efficiency with a comparable number of Gaussian primitives. Overall, these results demonstrate the suitability of Mobile-4DGS for resource-constrained mobile deployment, offering an effective trade-off among visual quality, rendering speed, storage cost, and training efficiency.

\noindent\textbf{Qualitative Results:}
Fig.~\ref{fig:nvs} provides a qualitative comparison of Mobile-4DGS against advanced methods, including 3DGS~\cite{kerbl2023gaussiansplatting}, Speedy-Splat~\cite{hanson2024speedy}, and Mobile-GS~\cite{mobilegs}. While 3DGS achieves high fidelity, its massive Gaussian count prohibits mobile deployment. Conversely, Speedy-Splat reduces primitives but introduces significant blurring in complex regions. Mobile-4DGS effectively bridges this gap. 
By optimizing the Gaussian distribution, our method preserves high-frequency details and sharp structural edges, such as mechanical components and textures, with greater clarity than previous methods.
The rendered results of Real-Time4DGS~\cite{real-time4dgs}, OMG4~\cite{omg4}, and our proposed Mobile-4DGS are compared with the corresponding ground-truth images, as depicted in Fig.~\ref{fig:nvs_4d}. Enlarged regions highlight challenging dynamic content, including rapid hand motion, object interactions, and fine structural details. Mobile-4DGS produces visually faithful results with sharper boundaries and fewer motion-related artifacts, while closely preserving the appearance of the ground truth. Moreover, it achieves this rendering quality using substantially fewer Gaussian primitives and considerably less storage than Real-Time4DGS, while also providing a more compact representation than OMG4.
 Across all of these scenarios, Mobile-4DGS consistently utilizes fewer primitives to approximate Ground Truth, yielding minimal storage requirements and enabling high-quality mobile rendering.

\begin{table*}[t!]
\centering
\caption{\textbf{Ablation study of our Mobile-4DGS on the Mip-NeRF 360 dataset.} 
MC-SEA denotes the proposed Monte Carlo Specular Energy Aggregator. 
$\Delta c$ represents the SH offset predicted by our Attribute-Conditioned SH Enhancement module. 
We also ablate the proposed Multi-view Alpha-based Densification and Pruning.
}

\scriptsize
\def\arraystretch{1.2}
\resizebox{0.8\linewidth}{!}
{
\begin{tabular}{l|cccccc}
\toprule
Method & PSNR $\uparrow$ & Storage (MB) $\downarrow$                     &Peak VRAM$\downarrow$& FPS$\uparrow$                               & \#Points $\times 10^6$ $\downarrow$          & Train (min) $\downarrow$                  \\ \hline
Mobile-4DGS& 27.16&\colorbox{green!30}{3.9}&\colorbox{green!30}{416}&\colorbox{green!30}{132}&\colorbox{green!30}{0.36} &  \colorbox{lime!30}{12}\\ \midrule
                  w/o MC-SEA  &                 26.72&                                               \colorbox{green!30}{3.9}&\colorbox{green!30}{416}&                                              \colorbox{green!30}{132}&                                              \colorbox{green!30}{0.36} &                                           \colorbox{lime!30}{12}\\
 w/o $\Delta c$ & 26.86& \colorbox{green!30}{3.9}& \colorbox{green!30}{416}& \colorbox{green!30}{132}& \colorbox{green!30}{0.36} & \colorbox{green!30}{10}\\
                  w/o Multi-view densify&                \colorbox{green!30}{27.21}&                                               18&1725&                                              21&                                              \colorbox{lime!30}{1.37}&                                           21\\ 
                  w/o Multi-view prune&                 \colorbox{lime!30}{27.19}&                                               \colorbox{lime!30}{6.2}&\colorbox{lime!30}{641}&                                              \colorbox{lime!30}{104}&                                              0.59&                                           15\\ \bottomrule
\end{tabular}}
\label{table:ablation}
\vspace{-3mm}
\end{table*}
\begin{figure*}[t!]
    \centering
    \includegraphics[width=0.99\linewidth]{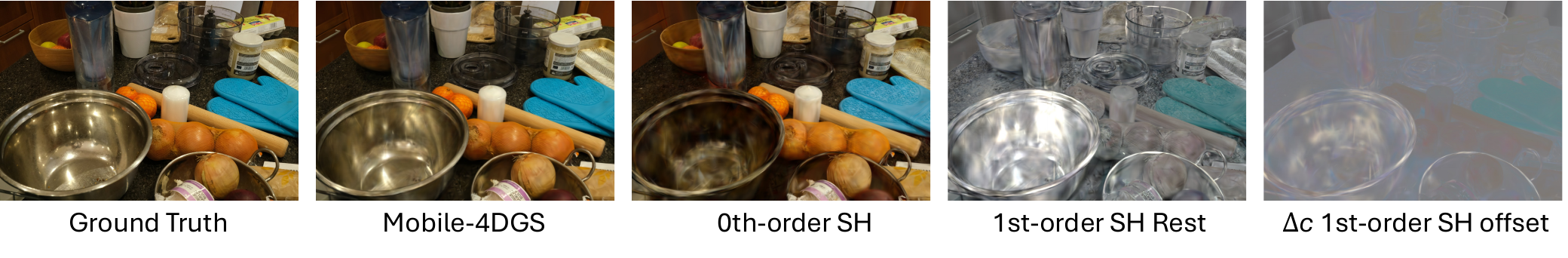}
    \caption{\textbf{Decomposition of the spherical harmonic components.}
The visual results are rendered using Mobile-4DGS under different spherical harmonic (SH) configurations. The comparison illustrates the contribution of different SH components to view-dependent appearance modeling and highlights their effects on color fidelity, fine-grained details, and overall rendering quality.}
    \label{fig:sh}
  \vspace{-4mm}
\end{figure*}

\subsection{Ablation Study}
We analyze the contribution of each component in Mobile-4DGS by progressively disabling these components, including Monte Carlo Specular Energy Aggregator (MC-SEA), enhanced SH offset $\Delta_c$, and Multi-view Alpha-based Densification and Pruning strategy. 
The full model achieves the best trade-off between reconstruction quality and efficiency, attaining competitive PSNR while requiring lower storage costs, fewer Gaussian points, and delivering the fastest FPS on mobile hardware. 
Removing MC-SEA or the SH offset degrades reconstruction quality while maintaining similar efficiency, indicating their positive and complementary contributions.
To ablate the proposed multi-view densification, we adopt the original single-view gradient-based densification~\cite{kerbl2023gaussiansplatting} for replacement.
Disabling multi-view densification substantially increases the number of Gaussians, memory usage, and storage cost.
It significantly reduces FPS, demonstrating the critical role of our proposed multi-view densification for compact Gaussian structure.
Similarly, removing multi-view pruning increases model size and reduces rendering speed, highlighting its importance in eliminating redundant primitives.
Overall, these results demonstrate that our proposed components contribute to achieving an improved trade-off between quality and efficiency, and their combination is essential for high-quality real-time mobile rendering.

\noindent\textbf{Analysis of Spherical Harmonic Decomposition:} 
Fig.~\ref{fig:sh} illustrates the decomposition of our Spherical Harmonic (SH) components, validating the efficacy of our enhanced first-order SH representation. We can find that the 0th-order SH in Mobile-4DGS primarily reconstructs the base diffuse color and global illumination of the scene, while the 1st-order and $\Delta$1st-order SH components are specialized to capture high-frequency structural variations and refine local context. 
These results demonstrate that our proposed first-order SH strategy effectively captures intricate textures and complex specular, enabling high-quality rendering that closely approximates the ground truth,  even under mobile hardware.

\noindent\textbf{Ablation of the Depth-Order Certificate.}
Table~\ref{tab:doc_ablation} evaluates the proposed depth-order certificate (DOC) under different reuse budgets. Separating static and dynamic primitives already reduces the ordering overhead compared with monolithic processing, while DOC further exploits temporal coherence by reusing previously committed depth orders. The exact setting ($r=0$) provides strict order preservation but is conservative, whereas increasing $r$ substantially improves the reuse rate and reduces both worker-side computation and index-buffer communication at the cost of a bounded depth-order deviation. Moreover, the higher reuse rate observed under denser frame requests demonstrates that DOC benefits naturally from stronger inter-frame temporal coherence. Overall, these results show that DOC provides a controllable trade-off between depth-order accuracy and rendering efficiency for dynamic mobile Gaussian Splatting.

\begin{table}[t]\centering\small
\caption{Ablation of the depth-order certificate (200k Gaussians, 8.1\% animated,
6\,s clip, 60\,Hz playback). \emph{ms/s} is worker time spent on
projection, sorting and merging. \emph{MB/s} is index-buffer upload. \emph{drift}
is the worst depth violation of the presented order in mean Gaussian radii.
}
\label{tab:doc_ablation}
\scriptsize
\def\arraystretch{1.2}
\resizebox{0.99\linewidth}{!}
{
\begin{tabular}{lccccc}
\toprule
Policy & reuse $\uparrow$ & ms/s $\downarrow$ & MB/s $\downarrow$& $\Delta$ms/s $\downarrow$& drift ($r$) \\
\midrule \midrule
Monolithic                 & 0.0\%        & 124.7 & 23.2 & $+53\%$      & 0.01 \\
Two-stream (separation)    & 0.0\%        & 81.3  & 23.2 & ---          & 0.01 \\
DOC certified ($r{=}0$)    & 0.6\%        & 104.1 & 23.1 & $+28\%$      & 0.01 \\
DOC $r{=}0.25$             & 65.5\%       & 25.8  & 8.0  & $-68\%$      & 0.26 \\
DOC $r{=}0.5$              & \colorbox{lime!30}{81.0\%}       & \colorbox{lime!30}{13.9}  & \colorbox{lime!30}{4.4}  & \colorbox{lime!30}{$-83\%$}      & 0.44 \\
DOC $r{=}1$                &\colorbox{green!30}{90.2\%}       & \colorbox{green!30}{7.2}   & \colorbox{green!30}{2.3}  & \colorbox{green!30}{$-91\%$}      & 1.28 \\
DOC $r{=}0.5$ @60\,Hz req. & \colorbox{green!30}{90.2\%}       & 14.7  & 4.5  & $-82\%$      & 0.64 \\
\bottomrule
\end{tabular}}
\vspace{-3mm}
\end{table}

\begin{figure}[t!]
    \centering
    \includegraphics[width=0.95\linewidth]{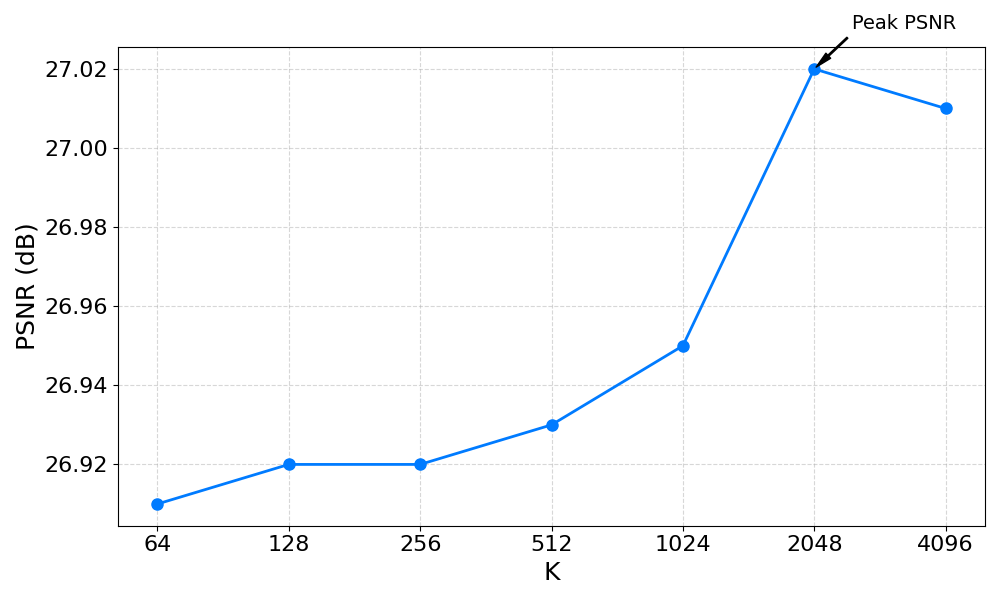}
    \caption{ \textbf{Impact of Monte Carlo Sampling Points $K$ on Reconstruction Quality.} We evaluate PSNR across varying sampling densities on the Mip-NeRF 360 dataset. Performance follows an upward trend as $K$ increases, with a notable saturation point appearing at $K = 2048$, indicating an optimal reconstruction fidelity.   }
    \label{fig:K}
    \vspace{-3mm}
\end{figure}

\subsection{Monte Carlo $K$ Sampling Point Analysis}

For sampling Density $K$, we investigate the sensitivity of our proposed method to the number of Monte Carlo sampling points $K \in \{64, \dots, 4096\}$. As illustrated in Fig.~\ref{fig:K}, the reconstruction quality PSNR demonstrates a monotonic improvement as $K$ increases from 64 to 2048. 
Specifically, we observe that increasing the density significantly reduces the variance inherent in the Monte Carlo estimator.
In particular, lower sampling rates ($K < 2048$) result in sub-optimal approximations of the integral.
Beyond this point ($K = 4096$), the gain in PSNR plateaus, suggesting that the estimator has converged.
Given the linear increase in computational overhead associated with larger $K$, we adopt $K =2048$ as the default configuration in our subsequent experiments to achieve a favorable rendering quality.

\begin{figure}[t!]
    \centering
    \includegraphics[width=0.95\linewidth]{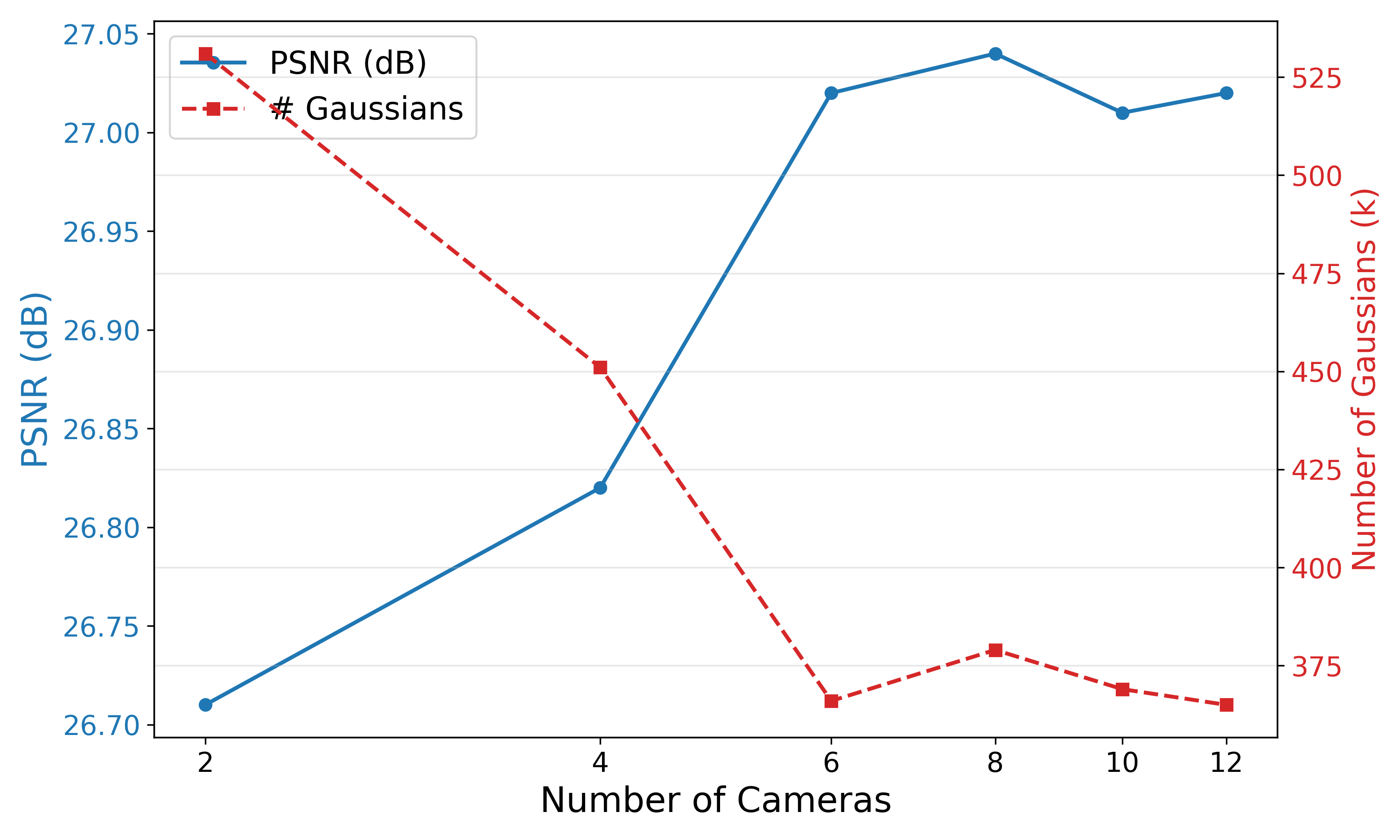}
    \caption{\textbf{Impact of the number of cameras in multi-view alpha-based densification.} We analyze the scaling impact of varying camera counts in our multi-view densification strategy on the Mip-NeRF 360 dataset. While PSNR steadily improves and stabilizes with additional views, the total number of Gaussian primitives decreases significantly, demonstrating the effectiveness of our densification in avoiding aggressive densification while maintaining high-fidelity reconstruction.}
    \label{fig:camera_count}
    \vspace{-4mm}
\end{figure}

\subsection{Camera Count in Multi-view Alpha-based Densification}

 To evaluate the robustness and efficiency of our multi-view densification module, we conduct an ablation study by varying the number of input camera views from 2 to 12. As illustrated in Fig.~\ref{fig:camera_count}, we observe a synergistic relationship between viewpoint coverage and representation efficiency. Specifically, increasing the camera count from 2 to 6 yields a significant PSNR improvement.
Crucially, as the number of views increases, the total Gaussian count drops significantly. This suggests that stronger multi-view constraints allow our densification strategy to more accurately localize geometric primitives for densification, effectively avoiding overfitting and redundancy. 
Therefore, we choose to sample 6 views for our multi-view densification, resulting in a superior trade-off between rendering performance and training burden.

\section{Conclusion}
In this work, we presented Mobile-4DGS, a substantial journal extension of Mobile-GS and Flux-GS that unifies compact static and dynamic Gaussian Splatting within a deployment-oriented explicit representation. The framework follows a common design principle: appearance corrections are decoded and baked before deployment, primitive growth is controlled through multi-view guidance, and temporal variation is evaluated using elementary per-Gaussian operations rather than a deformation network. Static scenes arise as the zero-motion special case of the same representation. Mobile-4DGS does not aim to dominate every unconstrained reconstruction metric; instead, it provides a favorable trade-off among visual fidelity, model size, training cost, and real-device throughput. The results demonstrate that continuous-time dynamic rendering can be deployed on mobile hardware using a model of only a few megabytes, while static scenes retain competitive quality and high rendering speed. We believe this unified design provides a practical basis for mobile novel-view rendering and volumetric video playback.

\bibliographystyle{ieeetr}
\bibliography{main}



\begin{IEEEbiographynophoto}{Xiaobiao Du}
is currently pursuing a Ph.D degree at the University of Technology Sydney, Australia. His research interests include static 3D reconstruction, 4D reconstruction, 3D object generation, and video generation. He is particularly interested in improving few-shot 3D reconstruction with generative prior.
\end{IEEEbiographynophoto}

\begin{IEEEbiographynophoto}{Zhen Fang} is the Australian DECRA Fellow, and he received the PhD degree in the Faculty of Engineering and Information Technology, University of Technology Sydney, Ultimo, Australia. He is a member of the Decision Systems and e-Service Intelligence (DeSI) Research Laboratory, Australian Artiﬁcial Intelligence Institute, University of Technology Sydney (UTS). Currently, he is the lecturer at UTS. His research interests include transfer learning and out-of-distribution learning. He has published over $60$ papers in top-tier conferences and journals, e.g., IEEE TPAMI, IEEE TNNLS, IEEE TCYB, ICML, NeurIPS, and ICLR. He received the NeurIPS 2022 Outstanding Paper Award, the 2023 Australasian AI Emerging Researcher Award and NeurIPS 2025 Top Area Chair.
\end{IEEEbiographynophoto}

\begin{IEEEbiographynophoto}
{Tianqing Zhu} received the B.Eng. and M.Eng.
degrees from Wuhan University, Wuhan, China in 2000 and 2004 respectively, and
the Ph.D. degree in computer science from Deakin University, Australia, in
2014. She is currently a Professor and the Dean with the Faculty of Data
Science at the City University of Macau. Before that, she was an Associate
Professor with the School of Computer Science, University of Technology Sydney,
and a Lecturer with the School of Information Technology, Deakin University,
from 2014 to 2018. Her research interests include privacy-preserving and AI
security. She has published more than 400 papers in refereed international
journals and refereed international conferences proceedings, including many
articles in IEEE Transactions and journals.
\end{IEEEbiographynophoto}

\begin{IEEEbiographynophoto}{Richard Hartley}(Fellow, IEEE) is a member of the computer vision group with the Research School of
Engineering, ANU, where he has been since January,
2001. He is also a member of the computer vision
research group in NICTA. He worked with the GE Research and Development Center from 1985 to 2001,
working first in VLSI design, and later in computer
vision. He became involved with Image Understanding and Scene Reconstruction working with GE’s
Simulation and Control Systems Division. He is an
author (with A. Zisserman) of the book Multiple View
Geometry in Computer Vision.
\end{IEEEbiographynophoto}

\begin{IEEEbiographynophoto}{Xin Yu}
received the BS degree in electronic engineering from the University of Electronic Science
and Technology of China, Chengdu, China, in 2009,
the PhD degree from the Department of Electronic
Engineering, Tsinghua University, Beijing, China,
in 2015, and the PhD degree from the College of
Engineering and Computer Science, Australian National University, Canberra, Australia, in 2019. He
is currently a senior lecturer with the University of
Queensland. His research interests include computer
vision and image processing.
\end{IEEEbiographynophoto}

\vfill

\end{document}